%% file: main.tex
\documentclass{article}

\usepackage{PRIMEarxiv}
\usepackage[utf8]{inputenc}
\usepackage[T1]{fontenc}
\usepackage{url}
\usepackage{booktabs}
\usepackage{amsfonts}
\usepackage{microtype}
\usepackage{fancyhdr}
\usepackage{graphicx}
\usepackage{enumitem}
\usepackage[section]{placeins}  % keep floats inside their own section
\input{math_commands}
\usepackage{hyperref}
\hypersetup{pdfborder={0 0 0},
  pdftitle={Co-Sight Harness: Externalized Control Assets and Deterministic
            Enforcement for Long-Horizon Business Agents},
  pdfauthor={Luyao Wang, Xijun Zhu, Zhaohui Hao, Xiwen Zhang, Yanqin Gao, Yongsheng Du,
             Lingjun Huang, Shaowei Zhao, Gang Li, Qifan Yu, Baoli Wang}}

\graphicspath{{figures/}}
\title{Co-Sight Harness: Externalized Control Assets and Deterministic
       Enforcement for Long-Horizon Business Agents}

\author{
  \normalfont
  Luyao Wang, Xijun Zhu, Zhaohui Hao, Xiwen Zhang$^{\ast}$, Xingyu Wen,\\
  Weishan Li, Yuyang Liao, Yanqin Gao, Yongsheng Du, Lingjun Huang,\\
  Shaowei Zhao, Gang Li, Qifan Yu, Baoli Wang\\[3pt]
  \small ZTE Corporation\\[2pt]
  \small Email: \url{AIM@zte.com.cn}
}

\begin{document}

\noindent\includegraphics[width=3.5cm]{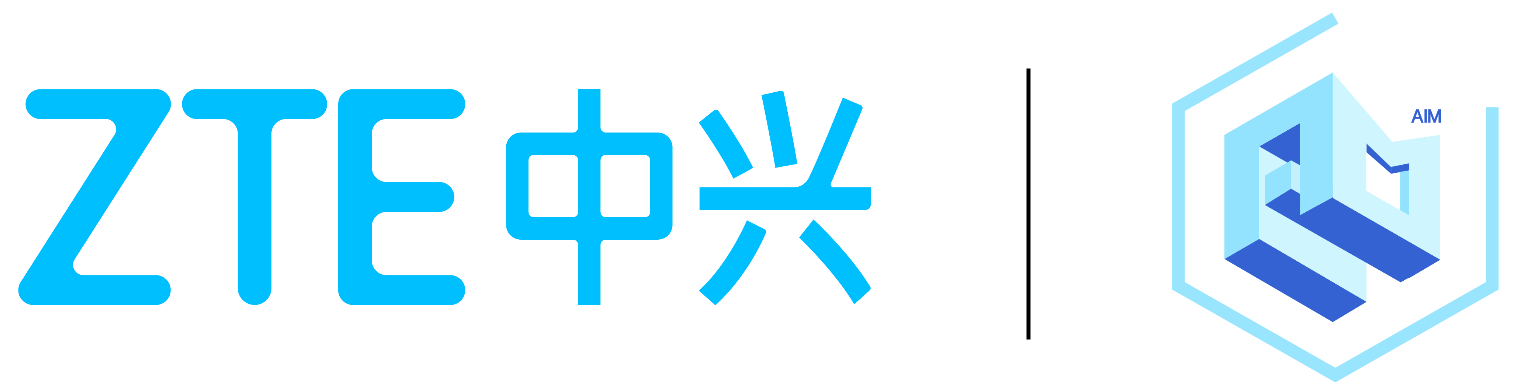}\par
\vspace{0.4cm}
\maketitle

\begin{abstract}
Long-horizon business workflows fail in two ways: errors accumulate as a
chain of activities lengthens, and the knowledge an activity needs is
written in a different vocabulary from the user's request. We study whether
such workflows can be made reliable without changing the language model.
Our method keeps the backbone fixed and moves the workflow definition,
state and domain knowledge into versioned control assets outside the model.
A state machine and machine-checkable acceptance gates enforce these assets
during execution; an independent auditor grades the resulting trace. The
assets are updated only after a regression check and human approval.We evaluate the method on three deployed workflows exceeding 200 steps.
On Domain A, a 120-case end-to-end evaluation set shows unattended
completion rising from 50\% to 100\%, retrieval recall from 56\% to
91.67\%, and end-to-end latency dropping from 440 to 159 minutes, with
scenario-analysis recall rising from 45\% to 93\%. Two other domains were
taken over by teams that did not write the assets, one in four days and one
in seven days. A controlled ablation separates the two failure sources. The
code is available at
\url{https://github.com/ZTE-AICloud/Co-Sight/tree/master/Co-Sight%20Harness}.
\end{abstract}

\input{introduction}
\input{related_work}
\input{problem_setting}
\input{method}
\input{experiments}
\input{conclusion}

\clearpage
\bibliographystyle{plainnat}
\bibliography{references}

\clearpage
\appendix
\input{appendix}
\input{statements}

\end{document}

%% file: math_commands.tex
\usepackage{amsmath,amssymb}
\usepackage{booktabs,multirow,array,tabularx}
\usepackage{algorithm,algpseudocode}
\usepackage{enumitem}
\usepackage{natbib}
\usepackage{graphicx}
\usepackage{url}
\usepackage{caption}
\usepackage{pifont}

\newcommand{\CR}{\mathrm{CR}}
\newcommand{\CoP}{\mathrm{CoP}}

\newcommand{\HI}{\mathrm{HI}}
\newcommand{\InvViol}{\operatorname{InvViol}}
\newcommand{\Reg}{\operatorname{Reg}}

\newcommand{\Compile}{\operatorname{Compile}}
\newcommand{\Execute}{\operatorname{Execute}}

\newcommand{\Fone}{\mathrm{F1}}
\newcommand{\GoldEval}{\textsc{GoldEval}}

\newcommand{\Eval}{\mathrm{Eval}}
\newcommand{\Approve}{\operatorname{Approve}}
\newcommand{\Gate}{\operatorname{Gate}}
\newcommand{\trig}{\operatorname{trig}}
\newcommand{\Mtheta}{M_\theta}
\newcommand{\Rcal}{\mathcal{R}}
\newcommand{\Treg}{\mathcal{T}_{\mathrm{reg}}}
\newcommand{\Lctl}{L_{\mathrm{ctl}}}

\newcommand{\Rrun}{R_{\mathrm{run}}}
\newcommand{\Rbuild}{R_{\mathrm{build}}}

\newcolumntype{Y}{>{\raggedright\arraybackslash}X}

%% file: introduction.tex
\section{Introduction}
\label{sec:introduction}

\subsection{Background and Motivation}
\label{sec:intro:motivation}

An agent harness is the software layer that mediates between a large
language model (LLM) and its environment: it decides what context the
model sees, which tools may be called, how state is persisted, and when an
output is accepted. Most harness optimization to date has been developed on
short-horizon coding tasks, where a compiler, a test suite and a
continuous-integration loop supply a cheap, timely and unambiguous success
signal. Under that regime a harness can be searched, rewritten or
self-modified, because a candidate can be scored in minutes.

Knowledge-intensive business workflows have no such signal. A
requirements-to-design chain in telecommunications spans more than 200
serial activities over hours or days, produces documents rather than
programs, and is judged by domain experts against acceptance criteria that
are only partly formalizable. There is no executable oracle for whether a
scenario analysis is faithful to the requirement, and a failure at activity
$i$ is often not observable until activity $j\gg i$ consumes its output.
Consequently, techniques that improve pass rate on executable benchmarks do
not transfer directly: neither the reward signal nor the failure locality
assumptions hold.

In a 200+ step requirements-to-design workflow, we measured a 50\%
unattended completion rate, 68\% generation recall and 56\%
average retrieval recall on a 120-case end-to-end evaluation set, before
any intervention; a single requirement took more than two hours and its
output was not directly usable. All percentages reported in this paper
are measured on this evaluation set (Section~\ref{sec:experiments});
the descriptive C-grade observations are reported separately.
These failures have two parallel causes. First, errors compound along a serial chain: under
an independence assumption the end-to-end success probability is the
product of per-step reliabilities, so a step-wise reliability of 0.95 over
ten steps already gives $0.95^{10}\!\approx\!60\%$. Second, the input side
suffers a semantic gap: user requests are phrased in business language
while the knowledge base is written in development language, so direct
vector retrieval misses the evidence that each activity needs. A purely
probabilistic fix---a stronger model, a longer prompt, or more retries---
changes neither the product structure of the failure nor the missing
evidence. We refer to these as the \emph{control-structure line} and the
\emph{knowledge-supply line}, and treat both as properties of the harness
rather than of the model.

\subsection{Problem Statement and Approach}
\label{sec:intro:question}

\textbf{With the backbone model held fixed, can a fixed set of control
invariants together with evidence-driven updates to versioned control
assets make long-horizon business workflows both completable and
transferable across teams?} We study this question by treating the harness
as an \emph{artifact}: the control data that governs a run is externalized
into versioned assets, enforced by deterministic code, and audited
read-only, while the model remains a frozen inference component. Fixing
$\Mtheta$ is deliberate: it removes model scaling as a confound, so the
reported differences cannot be explained by a change in underlying model
capability. \emph{Fixed} refers to the backbone; the invariant set is
explicitly declared, versioned with the assets, and shared across domains,
while $K$ is revised under evidence and human approval so the method adapts
its control data without learning or rewriting the harness.

We propose a model-fixed method with three invariants over a frozen
backbone $\Mtheta$. (1) An utterance $u$ is compiled into versioned control
assets $K$ with $K\cap\theta=\varnothing$; the compiler validates hard
checks and quality gates before $K$ is admitted, and an independent
verifier grades the artifact without seeing the generator's rationale.
(2) A declarative workflow $W$, a state machine $\Sigma$, hooks and
human-in-the-loop gates enforce $K$'s invariants; illegal transitions are
blocked, and a rejected stage triggers evidenced rework under a bounded
retry budget. (3) A read-only audit maps the trajectory $\tau$ to evidence
$z$, and asset updates $K_{t+1}$ are admitted only after a regression gate
and human approval. The four control planes, dual knowledge engines,
golden-set evaluation and confidence grading are \emph{deployment
instantiations} of this method in business domains, not additional
mechanisms of the method itself; Figure~\ref{fig:architecture} shows the
architecture and its three isolation boundaries. Section~\ref{sec:method}
gives the formalization and a staged algorithm.

\subsection{Contributions}
\label{sec:intro:contributions}

On Domain A, a 120-case end-to-end evaluation set shows unattended
completion rising from 50\% to 100\%, retrieval recall from 56\% to
91.67\%, and end-to-end latency dropping from 440 to 159 minutes, with
scenario-analysis recall rising from 45\% to 93\%. The same control assets
are transferred to two further domains by non-original teams in 2
person$\times$4 days and 1 person$\times$7 days. The comparison holds the
backbone fixed and reports outcomes on the 120-case evaluation set across
three deployed domains.

We associate these gains with per-activity gates with bounded rework,
externalized state, generation--evaluation separation, and externally
supplied domain knowledge, rather than with a change in model capability or
longer prompts. Unlike methods that rewrite or search the harness itself,
our method fixes the harness as the object under study and asks which
invariants it must enforce. All comparisons hold the backbone fixed and
report outcomes on the evaluation set across the three deployed domains. Our
contributions are:

\begin{itemize}[leftmargin=1.6em,labelsep=0.6em,itemsep=2pt,label=\textbullet]
\item \textbf{Method.} We describe a method that keeps the backbone fixed
  and places control data in external, versioned assets. It has three
  invariants (Section~\ref{sec:problem}) and a staged procedure
  (Algorithm~\ref{alg:loop}). We release a reference implementation of the
  asset compiler, execution runtime and read-only auditor (Appendix~A).
\item \textbf{Field evidence.} A three-domain deployment study with
  a 120-case evaluation set and disjoint ability and regression subsets,
  covering end-to-end completion, content quality and cross-team
  transferability.
\item \textbf{Structured decomposition and attribution chain.} We
  separate the shared invariant set from per-domain configuration
  constants, organize the measurements along a control-structure line and a
  knowledge-engineering line, contrast them with a descriptive pre-asset
  baseline, and interpret non-monotonic tuning by the rollback branch of
  the asset-admission rule defined in Section~\ref{sec:method}. An
  attribution chain with evidence grades is given in
  Section~\ref{sec:attribution}.
\end{itemize}

The contribution is an explicitly declared invariant set maintained over
versioned control assets: we quantify both failure lines on deployed
workflows exceeding 200 steps, where neither had been quantified before,
and show the same invariant set shared by three deployed domains.
Section~\ref{sec:problem} defines the objects, the failure model and the
objective; Section~\ref{sec:method} states the method and its staged
algorithm; Section~\ref{sec:experiments} reports the three-domain field
measurements and a controlled ablation that isolates the two failure lines.

%% file: related_work.tex
\section{Related Work}
\label{sec:related}

We organize prior work along the two failure lines identified in
Section~\ref{sec:intro:motivation}. The first concerns where control
resides and how the model--harness pair should be evaluated; the second
concerns how knowledge reaches each activity, how process evidence is
produced, and when a human may safely exit the loop.

\subsection{Deterministic Control and Harness Evaluation}
\label{sec:related:control}

A growing line of work places a deterministic control plane around the
model rather than delegating governance to further LLM orchestration.
\citet{madatha2026controlplane} treats agent definitions as a managed
supply chain with content addressing, hash-chained audit logs and a phase
state machine, showing in a study of 10,008 public repositories that
10.1\% of tracked configuration paths are exact duplicates across
independent repositories and that fewer than 1\% declare permission
boundaries. \citet{fournier2026processharness} wraps a deterministic
workflow engine with a policy-governed agentic layer whose hooks contribute
reasoning at designated control points while the engine retains structural
authority. \citet{orzechowski2026sla} admits completion only after an
independent control plane observes a deadline-bound service-state
predicate, rather than trusting the agent's self-assessment;
\citet{zhao2026evoml}, \citet{gupta2026agenttrustci} and
\citet{shan2026statemirror} provide validation gates, build-time trajectory
tests and deterministic pre-execution checks respectively. Collectively
these results establish that compliance should be enforced structurally,
not requested probabilistically.

A complementary strand optimizes the harness itself and reframes the unit
of evaluation. \citet{zhang2025aflow} searches a graph of code-represented
workflows, \citet{zhang2026dgm} iteratively rewrites its own code, and
\citet{pan2026rho} selects difficult trajectory coresets for retrospective
optimization; \citet{gefen2026cleo} instead minimizes the tool surface of a
fixed self-hostable model. \citet{huyut2026harnessmodel} shows that the
same model's compliance can differ by 11.5 percentage points across
harnesses, which makes the model--harness pair, rather than the model
alone, the proper evaluation unit. \citet{pan2026harnesspro} is our main
comparator: it searches a space of harness variants under executable
grading and returns a three-tier Pareto set of pass rate and cost. Our
procedure does not search a harness space: it holds the backbone and one
explicitly declared invariant set fixed, and admits versions of the asset
$K$ only through a regression check and human approval.
Algorithm~\ref{alg:loop} is therefore a constructive pipeline rather than
an optimizer. The contrast is one of problem setting---document-producing
chains judged by domain experts have no executable oracle with which to
score a candidate harness---rather than a claim of algorithmic superiority:
HarnessPro's search is reproducible, whereas our contribution is a
formalized control contract and field measurements on 200+ step deployed
chains.

The difference is one of object and horizon. These studies optimize or
govern a harness on short tasks with executable feedback; we study one
harness as an open-source system, formalize the invariants it
enforces, and measure it on deployed chains exceeding 200 steps where no
compiler supplies a verdict.

\subsection{Knowledge Supply and Process Evidence}
\label{sec:related:knowledge}

Semantic gaps between abstract goals and technical documentation motivate
multi-step retrieval planning~\citep{fang2026toolqp} and typed intermediate
representations that compile user purpose into executable semantic
boundaries~\citep{chen2026purposeir}. Experience reuse is represented by
\citet{ouyang2026reasoningbank}, \citet{suzgun2026dynamiccheatsheet} and
\citet{pan2026contextpilot}; \citet{yehudai2026survey} identifies
fine-grained and scalable evaluation as an open gap. This literature
primarily addresses \emph{what} knowledge an agent should see, whereas our
knowledge-supply line also addresses \emph{where} that knowledge is
versioned so that it survives a change of team.

On the evidence side, \citet{guo2026agenteval} reports a 2.17$\times$ gain
in failure-detection recall from DAG-structured step-level evaluation and
$\kappa=0.84$ agreement with human experts;
\citet{charoosaei2026llmregtests} finds that 8--17\% of LLM-generated
regression tests enforce faulty behaviour, which motivates human-authored
golden sets and a strict separation of ability and regression evaluation.
\citet{grace2026demystifying} supplies the terminology for that
separation; expert-verified long-horizon benchmarks with annotated atomic
criteria, such as \citet{cao2026longbanana}, inform our checklist design,
while \citet{abichandani2026rsd} and \citet{shu2026escrow} address
trajectory-level consistency and guarded continual maintenance.
Human-in-the-loop governance is formalized by
\citet{breckenridge2026ihitl} and \citet{cheng2026threepillar}, and the
long-horizon delivery gap is documented by \citet{liu2026deliverbench},
\citet{wu2026mattoolbench}, \citet{wei2026slate},
\citet{deng2026swebenchpro}, \citet{liu2026bats} and
\citet{erol2026costofpass}.

\subsection{Positioning}
\label{sec:related:positioning}

Deterministic-control work places enforcement outside the model but rarely
couples it to externalized knowledge supply or cross-team replication;
harness-optimization work searches or rewrites the harness under an
executable reward and reports outcome-level pass rate and cost rather than
content quality on document-producing chains; knowledge and evaluation
work supplies retrieval planning, golden sets and step-level judgement,
yet generally treats the harness as fixed context instead of versioned
control data. Table~\ref{tab:comparison} summarizes these differences.

We study the joint regime: a fixed backbone, control structure and
knowledge supply changed together, compliance enforced deterministically,
evaluation split into ability and regression, and the resulting assets
handed to a team that did not write them, on a 200+ step deployed
workflow. Our evaluation in Section~\ref{sec:experiments} is designed
around exactly this regime. Existing work rarely couples externalized
knowledge supply with machine-judged process evidence and
confidence-graded human exit under these conditions; that coupling, and
the field measurement of its effect, is the contribution of this paper.
Our formulation treats the versioned asset $K$ as the only
mutable object, updated discretely under the regression-gated admission
rule of Section~\ref{sec:method}, so any adaptation remains inspectable
and auditable.

\begin{table}[!htbp]
\centering
\caption{Comparison of harness optimization and control approaches.
Knowledge supply and cross-team replication are the two dimensions on
which our setting differs from prior work.}
\label{tab:comparison}
\small
\setlength{\tabcolsep}{3pt}
\renewcommand{\arraystretch}{1.12}
\resizebox{\linewidth}{!}{%
\begin{tabular}{llllllll}
\toprule
Method & Goal & Control locus & Knowledge supply & Deterministic & Evaluation & Replication & Long-horizon \\
\midrule
AFlow / DGM \citep{zhang2025aflow,zhang2026dgm} & workflow graph & in-workflow & none & no & outcome & no & no \\
RHO / AHE & pass rate & in-harness & partial & partial & outcome & no & partial \\
HarnessPro \citep{pan2026harnesspro} & pass + cost & in-harness & none & partial & synthesised tests & no & no \\
DSPy / GEPA & prompt or program & in-prompt & partial & no & outcome & no & no \\
ReasoningBank / DC / BATS & single component & in-skill / memory & partial & no & outcome & no & no \\
\textbf{Ours} & completion, quality, transfer & external $K$ & dual engines & state machine + hooks & \GoldEval, ability/regression split & 3 deployed domains & 200+ steps \\
\bottomrule
\end{tabular}%
}
\par\vspace{2pt}
{\footnotesize\raggedright \textit{Note.} ``partial'' denotes a mechanism
present in some but not all of the compared systems; DC denotes Dynamic
Cheatsheet. Comparisons are qualitative and follow the discussion above.\par}
\end{table}

%% file: problem_setting.tex
\section{Problem Setting}
\label{sec:problem}

\subsection{Long-Horizon Workflows and Control Assets}
\label{sec:problem:task}

A long-horizon business workflow is a chain of activities
$A=(a_1,\ldots,a_n)$ with $n$ exceeding 200; an activity may be a scenario
analysis, a requirements decomposition or a design synthesis. A harness
$h=\langle W,\Sigma,G,V\rangle$ mediates a frozen backbone $\Mtheta$ and
its environment, loading versioned control assets $K$. Here $W$ is a
declarative workflow, $\Sigma$ a state machine with transition function
$\delta$, $G$ the set of DoD gates $g_i$, and $V$ the verifier
configuration. Each activity is characterized by a five-dimensional
diagnosis $d(a_i)\in\mathbb{R}^5$ (LLM difficulty, freedom, acceptance
difficulty, retrieval dependence, vertical-knowledge dependence); the
measured distribution in our deployment is 58\% high difficulty, 50\% high
retrieval dependence, 67\% high vertical-knowledge dependence and 58\%
medium-or-above acceptance difficulty.

Control data are externalized outside the conversation,
$K\cap\theta=\varnothing$, in four classes: goals and plans, state and
checkpoints, intermediate artifacts, and distilled experience. This
separation is the first invariant: the model may generate business
artifacts, but it may not rewrite the process control that judges them. A
semantic mapping $\varphi$ relates user business language to development
knowledge; $\Rrun$ denotes runtime retrieval and $\Rbuild$ production-time
knowledge construction. A trajectory is
$\tau=\Execute(K,x;\Mtheta,\Rcal)$ with events $e_t$ and states
$\sigma_t\in\Sigma$; each activity has a DoD gate whose contract is
checked by \texttt{output\_schema.artifacts} (path pattern, minimum count,
minimum line count), and a failed gate triggers evidenced rework under a
bounded budget $b_t\le3$.
Figure~\ref{fig:architecture} makes the separation between control data and
business artifacts explicit.

Four \emph{deployment instantiation} views organize how $K$ is used:
business (goal, constraint, acceptance, risk), execution (orchestrator,
loop, graph, state, checkpoint), asset (knowledge, memory, tool, skill,
policy) and evidence (trace, evaluation, review, version, rollback). They
are configuration perspectives, not software modules and not additional
mechanisms of the method; their concrete configurations are reported in
Section~\ref{sec:method}. The four views are the \emph{control planes} of
the deployment, a term used throughout for configuration perspectives.

\subsection{Metrics and Notation}
\label{sec:problem:metrics}

We evaluate completion, content quality and resource use. Content quality
is measured by recall, precision and their harmonic mean over
hit/extra/missing sentence classes; completion is the fraction of tasks
whose final artifact passes the acceptance predicate; resource use is
token cost and wall-clock latency of a trajectory. These outcome metrics
are reported per measurement slice in Section~\ref{sec:experiments}. Two further
process indicators are used only diagnostically: human interventions
$\HI(\tau)=\#\{\text{HITL blocks}\}$ and cost per pass
$\CoP(K)=\mathbb{E}_{\tau}[C(\tau)]/\CR(K)$. Content quality $Q$ is a
deployment-side expert score, and $\HI$ enters the human-effort budget.

\GoldEval computes $R$, $P$ and $\Fone$ by a sentence-splitting,
classification (hit / extra / missing) and scoring pipeline for
LLM-judged items, and by exact or containment matching for deterministic
items. Ability evaluation ($\mathrm{pass@}3\ge90\%$) and regression
evaluation ($\mathrm{pass}^3=100\%$) are kept strictly separate; both are
\emph{thresholds specified by the deployment}, not measured pass rates.
Outcome-first judgement accepts any trajectory that reaches the goal,
including one that skips an unusable input or takes a different route.
Three-level drill-down yields metrics $M^{(0)}$ (task), $M^{(1)}$
(milestone) and $M^{(2)}$ (step), with localization accuracy $\LAop@\ell$;
$S$ denotes an auxiliary process score, $\InvViol$ counts illegal
transitions and $\Reg$ regression coverage.

\subsection{Failure Model}
\label{sec:problem:failure}

Without per-activity control, a failure at activity $i$ propagates silently
to activity $i+1$: end-to-end success is the product of per-step
reliabilities, so a step-wise reliability of 0.95 over ten steps already
gives $0.95^{10}\!\approx\!60\%$. This is an intuition about serial
structure rather than a fitted model; it ignores inter-step correlation
and rework cost. The measured baseline is a 200+ step chain with 50\%
unattended completion, and Figure~\ref{fig:error} illustrates the effect.
Per-activity gates with bounded rework cut the chain by turning a latent
failure into an observed, repairable event at the activity where it
arises. Cross-team outcomes are reported directly in
Section~\ref{sec:transfer}.

\subsection{Objective and Challenges}
\label{sec:problem:objective}

The engineering objective is to improve completion, content quality and
recall while holding regression coverage non-decreasing, human effort
within a fixed budget and the backbone unchanged. Control structure and
knowledge supply are changed jointly under these constraints; the ability
and regression thresholds are deployment constants rather than derived
quantities. Algorithm~\ref{alg:loop} is the constructive procedure that
realizes this specification, which separates the setting from
Pareto-frontier selection over a single performance--cost trade-off.

A 30-day observation of the pre-asset configuration grounds the failure
model. In that period 67 commits by 11 authors over 23 days touched 804
files; about 40\% of Skill changes concerned retrieval logic (45\%) or
rule constraints (35\%); 62\% of changes were runtime data; and 7 of 12
review-rule files were empty and therefore auto-passed. The concentration of
changes in retrieval and rule data motivates C2, and the empty-rule
observation motivates C3: a rule file that exists but is empty is
indistinguishable from one that passed. Section~\ref{sec:evidence} turns
this into a requirement for deterministic machine judgement.

Four challenges structure the method. \textbf{C1}: error compounding,
unobservable intermediate failures and state that is not persisted;
addressed by the control-structure line of Section~\ref{sec:control}.
\textbf{C2}: a semantic gap that corrupts every downstream step (56\%
retrieval recall at baseline); addressed by the knowledge-supply line of
Section~\ref{sec:evidence}. \textbf{C3}: coarse evaluation in which
ability and regression are mixed, so pass rates cannot be attributed;
addressed by \GoldEval and three-level drill-down in
Section~\ref{sec:evidence}. \textbf{C4}: expert experience that is not
transferable across teams; addressed by regression-gated asset reuse in
Section~\ref{sec:evidence} and measured in Section~\ref{sec:transfer}.

Four notational distinctions are maintained throughout. Verdict grades
$G_{0..3}$ (crash, generation failure, quality failure, accepted) differ
from audit layers $\mathcal{L}_{0..3}$ (rules, structure, semantic,
cross-task); $\mathrm{pass@}3$ and $\mathrm{pass}^3$ have different
directions; the weighting vectors used for compilation quality $v$,
runtime scoring $u^{\rm rt}$ and the six-dimension audit $w$ are never
interchanged; and \GoldEval is the single name for the golden-set
evaluator.

%% file: method.tex
\section{Method: Externalized Control Assets and Deterministic Enforcement}
\label{sec:method}

This section describes an engineering system organized around four
concrete failure modes of 200+ step business workflows. \emph{(P1)} Errors
compound along the serial chain because a failed activity is not observable
before its output is consumed. The control-structure line splits a
monolithic agent into 11 atomic subagents under an
Orchestrator$\to$Worker$\to$Review hierarchy, separates generation from
evaluation, and places a mechanically checkable DoD gate at every activity
exit; state, checkpoints and control data live outside the conversation, a
state machine admits only declared transitions, and a default-deny write
allowlist prevents the model from rewriting process control. \emph{(P2)}
User requests are phrased in business language while the knowledge base is
written in development language, so activities lack the evidence they
need. The knowledge-engineering line supplies four runtime retrieval
paths---semantic alignment, historical experience mining, agentic
full-text search and a unified retrieval agent---plus production-time
construction that back-fills knowledge. \emph{(P3)} Mixing ability and
regression judgements makes pass rates impossible to attribute. \GoldEval
golden sets, disjoint ability/regression evaluation, read-only trajectory
auditing and three-level drill-down localize failures; the observation
that 7 of 12 review-rule files were empty and therefore auto-passed is the
empirical reason for deterministic machine judgement. \emph{(P4)} Expert
experience does not survive a change of team. Control data is therefore
externalized as versioned assets, so a new team re-instantiates
configuration constants rather than code, and every asset update passes a
regression check and human approval.

The method maintains an explicitly declared invariant set over these
versioned assets with a frozen backbone $\Mtheta$: externalized control
assets, deterministic enforcement and a read-only evidence loop. It is
realized by an \emph{asset compiler}, a \emph{declarative execution
runtime} and a \emph{read-only trajectory auditor}. Four control planes,
dual knowledge engines, \GoldEval golden sets and confidence grading are
\emph{deployment instantiations}, reported at the end of this section;
\emph{field evidence} is given in Section~\ref{sec:experiments}. Release
identifiers and reproduction commands are collected in Appendix~A and the
Reproducibility Statement.

We call the rule that admits a new asset version $K'$ only after a
regression check on the disjoint regression set and explicit human
approval the \emph{regression-gated admission rule}; its rollback branch
reverts to the previous asset version when the check fails.

\begin{figure}[t]
\centering
\includegraphics[width=0.94\linewidth]{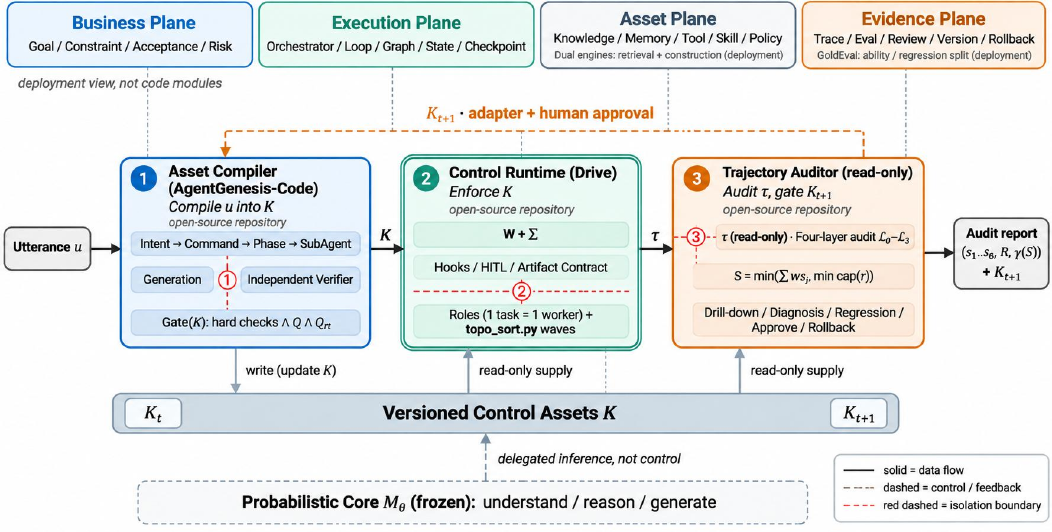}
\caption{Overview of the proposed method and its implementation.}
\label{fig:architecture}
\end{figure}

Figure~\ref{fig:architecture} reads as follows. An utterance $u$ is
compiled into versioned control assets $K$ (1); $K$'s invariants are
enforced by a declarative workflow $W$, a state machine $\Sigma$, hooks and
gates (2); and read-only trajectory auditing produces evidence with
regression-gated asset updates (3). The four control planes shown above the
three panels are deployment instantiations of this method, not software
modules; dual knowledge engines and \GoldEval evaluation are
deployment-level configuration. Three isolation boundaries---\ding{172}
generation vs.\ evaluation, \ding{173} control data vs.\ business
artifacts, \ding{174} read-only auditing---make compliance structurally
enforceable rather than model-dependent. Asset updates require an adapter
and human approval; the backbone $\Mtheta$ remains frozen throughout.

\subsection{Overview and Formalization}
\label{sec:formalization}

An utterance $u$ is compiled into versioned control assets $K$; the
candidate is admitted only when all fifteen boolean hard checks pass and
the compilation and runtime quality scores clear their deployment
thresholds ($Q\ge4.0$, $Q_{\rm rt}\ge3.5$). The compiler runs a six-phase
pipeline with intent routing, plan-mode approval, static verification and
an \emph{independent} verifier that grades the artifact without access to
the generator's rationale (isolation boundary~\ding{172}), which prevents a
model from grading its own output. Five-dimensional diagnosis $d(a_i)$
decides where to invest: activities with high retrieval or
vertical-knowledge dependence are prioritized, trivially checkable ones
are scripted, and the 11-subagent split granularity follows the grades.

Execution advances only through declared state transitions. Any
transition outside the declared set is rejected before it can mutate
state, fail-closed envelopes wrap each step, and every event is appended
to an audit log; rework is bounded by a deployment budget. The third
invariant maps the trajectory to evidence through four audit layers
$\mathcal{L}_{0..3}$ with an evidence cap, dimension scoring, red lines
and three-level drill-down (Section~\ref{sec:evidence}).

Algorithm~\ref{alg:loop} states the complete loop in five phases. Phase~I
externalizes intent into control data; Phase~II executes under
deterministic control; Phase~III derives evidence without mutating the
trace; Phase~IV separates ability from regression; Phase~V admits or rolls
back an asset version. The auditor is read-only and every merge requires
human approval.

\begin{algorithm}[ht]
\caption{Model-fixed control-asset loop}
\label{alg:loop}
\begin{algorithmic}
\Require utterance $u$, activities $A$, diagnosis $d(\cdot)$, frozen
  $\Mtheta$, retrieval $\Rcal$, regression set $\Treg$, budget $B$
\Ensure admitted assets $K_T$, evidence records $\{\Eval_t\}$
\Statex \textbf{Phase I: diagnosis and asset compilation}
\State Compute five-dimensional diagnosis $d(a_i)$ for every $a_i\in A$
\State Rank activities by retrieval and vertical-knowledge dependence;
  derive subagent split and context-window grade
\State Route utterance intent and emit clarification requests if
  the specification is underspecified
\State Assemble goals, plans, state schema, acceptance criteria and a
  risk register into candidate $K$
\State Run static verification over the assembled candidate
\State Grade the candidate with an independent verifier that receives
  the artifact but not the generator's rationale
\State Admit $K_0$ only if $\Gate(K_0)=1$ by the compilation gate;
  otherwise return to assembly with the failing checks
\Statex \textbf{Phase II: deterministic execution}
\For{each wave of the dependency-ordered DAG over $A$}
  \For{each activity $a_i$ in the wave}
    \State Load $K_t$, initialize $\sigma_0$ and an append-only event log
    \State Execute $a_i$ with $\Mtheta$ and $\Rcal$; record events $e_t$
    \State Apply $\delta(\sigma_t,e_t)$; reject transitions outside
      $\mathcal{T}(\sigma_t)$ by the state-machine invariant
    \State Fire pre- and post-execution hooks and enforce the DoD
      artifact contract
    \If{DoD gate $g_i$ fails}
      \State Attach the failing contract fields to the event log and
        trigger evidenced rework with $b_t\le r_{\max}$
      \State Require $\InvViol(\tau)=0$ before the reworked activity
        is admitted
    \EndIf
    \State Persist artifacts, state and checkpoint outside the
      conversation; enforce the default-deny write allowlist
  \EndFor
  \State Reconcile the wave barrier; hold high-risk actions at the
    double HITL gate
\EndFor
\Statex \textbf{Phase III: read-only evidence generation}
\State Map $\tau$ to $z$ through audit layers $\mathcal{L}_{0..3}$;
  never modify $\tau$, the plan or the artifacts
\State Apply evidence caps, red lines and dimension-wise minima
  (\eqref{eq:evidence-cap}--\eqref{eq:audit-total})
\State Drill down $M^{(0)}\!\to\!M^{(1)}\!\to\!M^{(2)}$ and compute
  $\LAop@\ell$ by \eqref{eq:drilldown}
\Statex \textbf{Phase IV: evaluation and gap formulation}
\State Score LLM-judged and deterministic items separately in \GoldEval
\State Report ability and regression outcomes independently over
  $\Treg$; never pool the two sets
\State Formulate a repair proposal $K'$ from localized gaps
\Statex \textbf{Phase V: gated admission}
\If{$\Reg(K')\ge\Reg(K_t)\ \wedge\ \Delta S\ge0\ \wedge\ \Approve(K')=1$}
  \State Set $K_{t+1}\gets K'$; commit the version and evidence
    record $\Eval_t$
\Else
  \State Set $K_{t+1}\gets K_t$ (rollback) and retain $\Eval_t$
    for audit
\EndIf
\State \Return admitted assets $K_T$ and evidence records
  $\{\Eval_t\}$
\end{algorithmic}
\end{algorithm}

The phase structure is what makes the method testable. Phase~I moves
decisions out of the conversation before any model call is charged to the
run; Phase~II makes every state change observable and every illegal change
impossible; Phase~III guarantees that evidence is a pure function of the
trace, so two auditors must reach the same verdict; Phase~IV prevents a
rising ability score from masking a regression; and Phase~V limits the
damage from a bad asset version. Each phase corresponds to a concrete
invariant (the compilation gate, the state-machine invariant,
the deterministic audit layers and the regression-gated admission rule) rather than to a
software module.
\FloatBarrier   % keep Algorithm~\ref{alg:loop} inside this subsection

\subsection{Deterministic Enforcement}
\label{sec:control}

The control-structure line $\Lctl$ suppresses error compounding. A
monolithic agent is split into 11 atomic subagents, each carrying a single
vertical domain; an Orchestrator$\to$Worker$\to$Review hierarchy separates
generation from evaluation; context windows are graded by task type
(4--8k for atomic operations, 12--16k for single-file edits, 20--32k for
cross-file analysis); eight classes of review rules cover format and
business checks; and a DoD gate admits only conforming artifacts, with
evidenced rework otherwise. This produced about $+10\%$ accuracy in the
deployment. This is the locality mechanism: a gate at activity $a_i$
converts a latent failure into an observed, repairable event before
$a_{i+1}$ can consume it, so the failure cannot propagate silently along
the serial chain as it does in a monolithic baseline.

Execution follows five stages, \textsc{Start}$\to$\textsc{Analysis}%
$\to$\textsc{Plan}$\to$\textsc{Executor}$\to$\textsc{Evaluation}, with
failure returning to the start. All four classes of control data are
persisted outside the conversation; state files are locked, the state
machine permits only legal transitions, and workflow directories are
written through a default-deny allowlist---an agent may modify business
artifacts but not process control (isolation boundary~\ding{173}). Every
transition appends an audit log; high-risk actions pass a double HITL gate;
and dependency-ordered DAG waves are dispatched with barrier
reconciliation and bounded retries $b_t\le3$. The implementation exposes
workflow dataclasses, inferred legal-transition inference, a state
transition guard hook, fail-closed envelopes and explicit
verdict-required gates. The plugin manifest states that no runtime engine
is bundled---the host runtime executes, while the plugin supplies control.
Two design principles follow from this separation: orchestration fixes
structure, and policy is data rather than code. We describe the code as it
is: the accountant role was removed and its duties moved to a deterministic
topological-sort helper.

\subsection{Knowledge Supply and Auditable Evidence}
\label{sec:evidence}

\paragraph{Knowledge-supply line (addresses C2).} Four knowledge gaps are
observed: retrieval limits, a semantic gap between user and development
language, imprecise definitions and missing navigation. Runtime retrieval
$\Rrun$ uses four paths: semantic alignment mapping $\varphi$ with
cold-start construction and dynamic update, historical experience mining,
agentic full-text search, and a unified retrieval agent. Production
construction $\Rbuild$ mines processes, builds knowledge and back-fills
automatically. These are deployment instantiations described with the
architecture/code distinction. On the 120-case evaluation set, recall rises
$45\%\!\to\!93\%$ for scenario analysis, $49\%\!\to\!92\%$ for
feature-domain change and $74\%\!\to\!90\%$ for subsystem change. Supplying
the evidence an activity actually needs raises golden-set recall by
16--48 percentage points across the three activities, which is the
input-side counterpart of the control-structure gain.

\paragraph{Measurable evidence.} \GoldEval maintains a 120-case golden
set with disjoint ability and regression subsets, each case with a
reference answer and a decision rule. LLM-judged items use
sentence splitting and hit/extra/missing classification to produce $R$,
$P$, $\Fone$; deterministic items use exact and containment matching.
Ability and regression evaluation are separated because mixing them makes
the pass rate unattributable; \citet{charoosaei2026llmregtests} shows that
LLM-generated regression tests can preserve faulty behaviour. Outcome-first
judgement accepts any valid route to the goal. Audit scoring uses an
evidence cap and a dimension-wise minimum rather than averaging:
\begin{equation}\label{eq:evidence-cap}
\mathrm{cap}(\varepsilon)=\{\textit{missing}\!:0,\ \textit{unobserved}\!:3,\
 \textit{present}\!:3,\ \textit{exercised}\!:5,\
 \textit{outcome-supported}\!:5\}.
\end{equation}
\begin{equation}\label{eq:dim-score}
s_j^{\rm raw}=f_j(\mathbf{b}_{\mathcal{C}_j})\in\{0,\ldots,5\},\qquad
s_j=\min\Bigl(s_j^{\rm raw},
 \min_{c\in\mathcal{C}_j}\mathrm{cap}\bigl(\varepsilon(c)\bigr)\Bigr).
\end{equation}
\begin{equation}\label{eq:audit-total}
R(\tau)=\{r:\trig(r,\tau)=1\},\qquad
S(\tau)=\min\Bigl(\sum_{j=1}^{6}w_j s_j(\tau),
 \min_{r\in R(\tau)}\mathrm{cap}(r)\Bigr).
\end{equation}
A triggered red line caps the total score and cannot be averaged away by
high dimension scores. The deterministic audit layers
($\mathcal{L}_0,\mathcal{L}_1,\mathcal{L}_3$) depend only on the
trajectory, which makes machine judgement reproducible. Three-level
drill-down and localization accuracy are
\begin{equation}\label{eq:drilldown}
M^{(0)}_{\rm task}\to M^{(1)}_{\rm milestone}\to M^{(2)}_{\rm step},\qquad
\LAop@\ell=\frac{\#\{\text{failures localized at granularity }\ell\}}
{\#\{\text{failures}\}} .
\end{equation}
The empty-rule observation of Section~\ref{sec:problem:objective} motivates
all of this: a rule file that exists but is empty is indistinguishable
from one that passed. Rule existence therefore carries no information
about whether the rule fired; deterministic machine judgement is
required.

\paragraph{Loop, confidence and replication.} The evaluation loop proceeds
from golden set to run trace, four metrics, root cause, repair proposal,
regression and version merge. An asset update is admitted only when
regression coverage does not decrease, the process score does not drop and
a human approves it; otherwise the previous version is retained
(rollback), as stated in Phase~V of Algorithm~\ref{alg:loop}.
Confidence is graded in three levels: low confidence requires human
confirmation, medium confidence uses LLM verification with sampling, and
high confidence uses deterministic checks with no LLM call. Humans may exit
the loop only when the DoD is mechanically decidable, the regression set
has sufficient coverage and historical judgements are stable; convergence
is not zero, and high-risk or unseen activity types always retain a human
backstop. The evaluation system evolved from seven generic LLM-scored
dimensions (v1.0, too coarse to localize), through expert rules compiled
into recall/precision formulas (v2.0, computable and reproducible), to
per-activity evaluation with stubbed workspaces and the previous step's
reference answer (v3.0, localizes to one activity). Externalized assets
also allow configuration changes instead of code changes: the 30-day
behaviour analysis described in Section~\ref{sec:mechanism} is the
descriptive counterfactual for a setting without control assets.

\paragraph{Shared invariants and per-domain constants.}
The method maintains an explicitly declared invariant set over versioned
assets and delegates only configuration constants to each domain.
Case-specific customization is thereby expressed
against one constrained interface---the versioned asset $K$ is the only
mutable object---rather than by re-authoring the control structure
for each request. Table~\ref{tab:invariants} enumerates what is shared and
what varies.

\begin{table}[!htbp]
\centering
\caption{Shared invariants versus per-domain configuration constants.}
\label{tab:invariants}
\small
\begin{tabularx}{\linewidth}{YY}
\toprule
Shared across domains (verbatim) & Per-domain configuration constants \\
\midrule
$K\cap\theta=\varnothing$; $\Compile(u)$ with $\Gate(K)$; $\delta$ with
$\InvViol=0$; generation--evaluation separation; read-only auditing;
ability/regression separation; regression-gated admission with human approval
& subagent split granularity (11); context-window grades (4--8k / 12--16k /
20--32k); contents of the 8 review-rule classes; golden-set contents
(120-case golden set); $\varphi$ lexicon; $Q$, $Q_{\rm rt}$, $r_{\max}$,
$\mathrm{pass@}3$, $\mathrm{pass}^3$; HITL high-risk list \\
\bottomrule
\end{tabularx}
\par\vspace{2pt}
{\footnotesize\raggedright \textit{Note.} The separation is exercised on
three deployed domains. The shared decision rule is the five-dimensional
diagnosis $d(a_i)$: it prioritizes activities with high retrieval or
vertical-knowledge dependence and scripts trivially checkable ones; only
the graded values change with the domain.\par}
\end{table}

\paragraph{Deployment instantiations.}
Table~\ref{tab:planes} maps the four control planes to their configurations
and per-plane measurements; \GoldEval, the task center, skill optimization
and the dual engines belong to this layer, whose capability descriptions
come from the field material and component documents. The asset compiler,
execution runtime and read-only auditor implement the compilation,
enforcement and auditing parts of the method and are publicly available; asset
admission follows the regression-gated admission rule with human approval. The
auditing plugin is
attached through an adapter calling \texttt{post-execute}.

\begin{table}[!htbp]
\centering
\caption{Deployment instantiations of the four control planes.}
\label{tab:planes}
\small
\begin{tabularx}{\linewidth}{lYY}
\toprule
Plane & Mechanism & Domain A measurement \\
\midrule
Business & Orchestrator$\to$Worker$\to$Review, generation/evaluation separation, 11 atomic subagents & about $+10\%$ accuracy \\
Execution & task-decomposed flow, externalized state, state machine + hooks, evidenced rework & completion $50\%\!\to\!100\%$; latency $-64\%$ \\
Asset & runtime retrieval (semantic alignment, experience mining, agentic full-text, unified agent) + production construction & retrieval recall $56\%\!\to\!91.67\%$; precision $36\%\!\to\!50\%$ \\
Evidence & iterative refinement, three-level drill-down, \GoldEval, ability/regression split & scenario recall $45\%\!\to\!93\%$; a further $+10\%$ accuracy \\
\bottomrule
\end{tabularx}
\par\vspace{2pt}
{\footnotesize\raggedright \textit{Note.} ``about'' marks approximate
figures as reported in the field material; all other values are as
measured.\par}
\end{table}

%% file: experiments.tex
\section{Evaluation on Three Deployed Domains}
\label{sec:experiments}

This section evaluates the method of Section~\ref{sec:method} on three
deployed 200+ step business domains. The evaluation is organized to answer
three questions in order: does the end-to-end chain complete; which of the
two failure lines accounts for the change; and does the effect survive a
change of team. These questions correspond to the challenges C1--C4 of
Section~\ref{sec:problem:objective}: completion under the first two
invariants tests C1, the two-line decomposition tests C2 and C3, and
replication by a non-original team tests C4. Each question is addressed
with a measurement slice of the 120-case evaluation set; its composition
and measurement boundaries are recorded in Appendix~G.

\subsection{Experimental Setup}
\label{sec:setup}

\paragraph{Domains and data.} Domain A is a telecommunications
requirements-to-design workflow (3 major phases, 39+ sub-skills, 200+
critical steps, 11 atomic activities); domain B is a software
requirements-to-design workflow; domain C is a multi-accelerator
model-adaptation workflow (framework adaptation, algorithm development,
precision and performance verification). The evaluation set contains 120
end-to-end cases with reference answers and decision rules, partitioned
into disjoint ability and regression subsets. It covers the four
full-flow activities, three retrieval sub-tasks, the three deployed
domains and the four-stage $\Fone$ values. All percentages in the result tables
are measured on this set; the activities, sub-tasks and
domains are measurement slices reported separately. Appendix~G records
the composition and the measurement boundaries.

\paragraph{Baseline.} The pre-intervention system is a monolithic agent with
a long prompt. The material gives per-item baselines: 50\% unattended
completion, 68\% generation recall with 67\% precision, 56\%
retrieval recall, and 169/114/440 minutes for requirements analysis,
subsystem design and the end-to-end chain. The baseline is the deployed pre-intervention system, which fixes the
comparison to a single production configuration; external systems are
discussed qualitatively. The after condition introduces orchestration,
retrieval and evaluation together, so the before/after difference is a
system-level comparison rather than an isolated manipulation; the
per-invariant readings of Section~\ref{sec:attribution} are therefore
associations measured on the evaluation set, not individually identified
causal effects.

\paragraph{Metrics and protocol.} Content quality and completion follow the
definitions in Section~\ref{sec:problem:metrics}; $R$, $P$, $\Fone$, $Q$,
$\CR$ and $L$ are the main-table columns, while $S$, $\InvViol$, $\HI$,
$\LAop@\ell$ and $\Reg$ are process indicators. $\mathrm{pass@}3$ and
$\mathrm{pass}^3$ are deployment thresholds distinct from measured pass
rates.

\paragraph{Measurement conditions.} The released components contain no
model weights: the backbone, embedding and reranking services are reached
through OpenAI-compatible API endpoints supplied by the deployment. One
backbone serves both the before and after conditions. The runtime is a
standard agent host with the three control plugins installed. \GoldEval
applies separate decision rules to LLM-judged and
deterministic items, and ability and regression sets are disjoint.
Component versions, reproduction entry points, the evaluation-set
composition, time windows and thresholds are listed in Appendices~A and~G.

\subsection{Main Results}
\label{sec:main_results}

On the 120-case evaluation set, unattended completion $\CR$ rises from 50\% to
100\%, generation recall from 68\% to 91.25\%, precision from 67\% to
84.38\%, retrieval recall from 56\% to 91.67\%, and end-to-end latency
$L(\tau)$ drops from 440 to 159 minutes. On the full-flow four
activities, measured recall is
78.08\%, 62.60\%, 67.93\% and 59.0\% against targets of 80\%, 60\%, 60\%
and 50\%; requirements analysis does \emph{not} reach its 80\% target.
The retrieval slice gives $56\%\!\to\!91.67\%$ recall and
$36\%\!\to\!50\%$ precision; the generation slice gives
$68\%\!\to\!91.25\%$ recall; cross-component collaboration
$\Fone$ $4\%\!\to\!61.4\%$ with latency $49.6\!\to\!24.7$ minutes.

\begin{table}[!htbp]
\centering
\caption{Domain A main results.}
\label{tab:main_results}
\small
\begin{tabularx}{\linewidth}{lYYY}
\toprule
\multicolumn{4}{l}{\emph{(a) completion and content quality}} \\
Metric & Before & After & Change \\
\midrule
Unattended completion & 50\% & 100\% & $+50$ pp \\
Generation recall & 68\% & 91.25\% & $+23.25$ pp \\
Generation precision & 67\% & 84.38\% & $+17.38$ pp \\
Retrieval recall & 56\% & 91.67\% & $+35.67$ pp \\
Requirements-analysis latency & 169 min & 56.4 min & $-67\%$ \\
Subsystem-design latency & 114 min & 26.5 min & $-77\%$ \\
End-to-end latency & 440 min & 159 min & $-64\%$ \\
\midrule
\multicolumn{4}{l}{\emph{(b) Full-flow recall against target}} \\
Activity & Measured & Target & Status \\
Requirements analysis & 78.08\% & 80\% & not reached \\
System-solution analysis & 62.60\% & 60\% & reached \\
Subsystem design & 67.93\% & 60\% & reached \\
Component design & 59.0\% & 50\% & reached \\
\midrule
\multicolumn{4}{l}{\emph{(c) activity-level recall and precision}} \\
Activity & Recall (before $\to$ after) & Precision (before $\to$ after) & $\Delta$ recall \\
Scenario analysis & 45\% $\to$ 93\% & 19\% $\to$ 44\% & $+48$ pp \\
Feature-domain change & 49\% $\to$ 92\% & 17\% $\to$ 37\% & $+43$ pp \\
Subsystem change & 74\% $\to$ 90\% & 72\% $\to$ 70\% & $+16$ pp \\
\midrule
\multicolumn{4}{l}{\emph{(d) Sub-task and cross-component outcomes}} \\
Slice & Metric & Before & After \\
Retrieval slice & recall & 56\% & 91.67\% \\
Retrieval slice & precision & 36\% & 50\% \\
Generation slice & recall & 68\% & 91.25\% \\
Cross-component collaboration & $\Fone$ & 4\% & 61.4\% \\
Cross-component collaboration & latency & 49.6 min & 24.7 min \\
\bottomrule
\end{tabularx}
\par\vspace{2pt}
{\footnotesize\raggedright \textit{Note.} ``pp'' denotes percentage
points (absolute difference) and ``\%'' denotes relative change. Sample
sizes are $n=120$ end-to-end cases throughout; (a)--(d) are measurement
slices of this set and are reported separately. Subsystem-change precision
moves from 72\% to 70\% while recall rises from 74\% to 90\%. The
$56\%\!\to\!91.67\%$ retrieval recall and the $68\%\!\to\!91.25\%$
generation recall each appear in two slices and are not pooled; values
retain the precision reported in the field material. For
the $120/120$ completions in (a), the Wilson 95\% interval is
$[0.969,1.000]$ and the Clopper--Pearson 95\% interval is
$[0.970,1.000]$; no interval is derived for the remaining aggregate
scores, which are averages over sentence-level judgements rather than
binomial counts.\par}
\end{table}

Table~\ref{tab:main_results} reports four measurement slices of the
120-case evaluation set: completion and content quality, the four
full-flow activities, activity-level recall and precision, and the
sub-task and cross-component outcomes. The backbone is held fixed
throughout. The four-stage $\Fone$ figures (62.66/60.50/62.27/56.4)
are recorded in the data card (Appendix~G).
Golden-set recall improves by 16--48 percentage points across the three
activities, while subsystem-change precision moves from 72\% to 70\%.

\begin{figure}[htbp]
\centering
\includegraphics[width=0.62\linewidth]{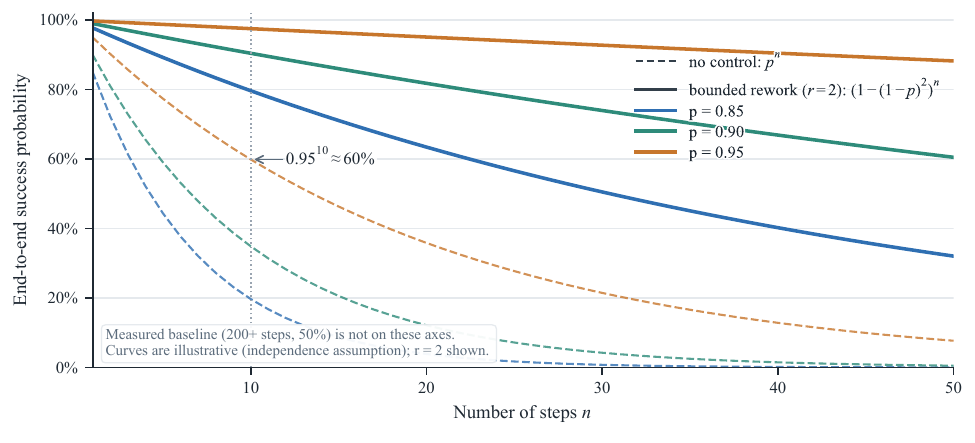}
\caption{Illustration of the product-form failure mechanism (purely
schematic; the measured 200+ step / 50\% baseline is reported in
Table~\ref{tab:main_results}(a) and is not a point on these curves).}
\label{fig:error}
\end{figure}

Figure~\ref{fig:error} supplies the mechanism sketch for these results.
Without external control, end-to-end success is the product of per-step
reliabilities; per-step gates with bounded rework give an upper bound under
an independence assumption. The curves are illustrative of this mechanism;
the measured 200+ step / 50\% baseline is reported in
Table~\ref{tab:main_results}(a) and is not a point on these curves.

\subsection{Mechanism Analysis}
\label{sec:mechanism}

Table~\ref{tab:planes} organizes the four deployment instantiations by
measured outcome: business about $+10\%$ accuracy; execution
$50\%\!\to\!100\%$ completion and $-64\%$ latency; asset
$56\%\!\to\!91.67\%$ retrieval recall; evidence $45\%\!\to\!93\%$
scenario recall and a further $+10\%$ accuracy. The five-dimensional
diagnosis is consistent with this ordering: activities with high retrieval
or vertical-knowledge dependence were addressed first, because they are
simultaneously the most error-prone and the most expensive to repair
downstream. The diagnosis of Section~\ref{sec:formalization} fixes this
ordering before execution (Section~\ref{sec:attribution}).

\begin{figure}[htbp]
\centering
\includegraphics[width=0.94\linewidth]{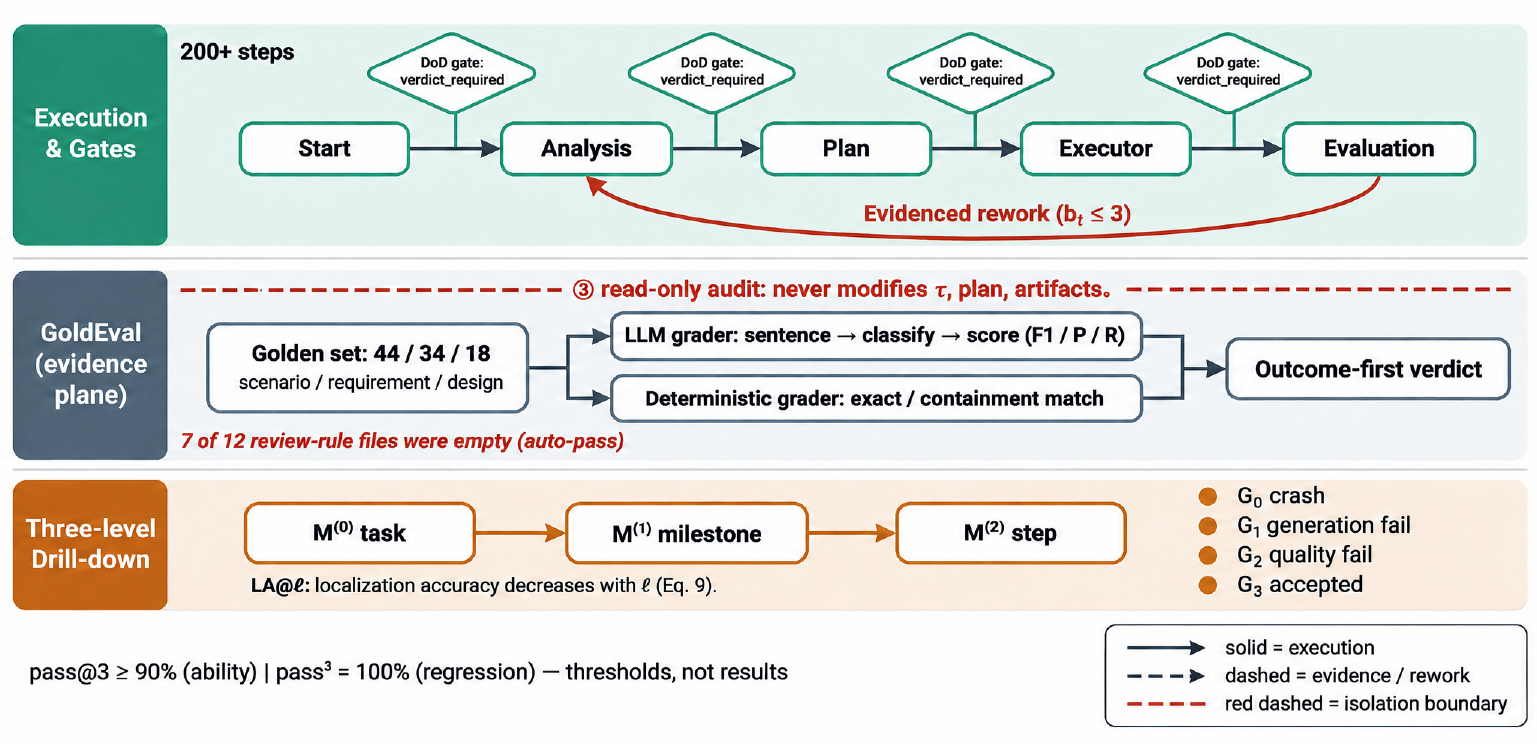}
\caption{Execution gating and evidence generation in the deployed system.
The gate labels and the bounded-rework label denote declared contracts; the
released runtime enforces the explicit verdict at the terminal state and
treats the rework budget as a design contract (Appendix~A). The golden
set contains 120 cases; $\mathrm{pass@}3$ and $\mathrm{pass}^3$ are
deployment thresholds, not measured pass rates.}
\label{fig:drilldown}
\end{figure}

Figure~\ref{fig:drilldown} shows the execution gates, the \GoldEval
pipeline and three-level drill-down that produce these measurements. Each
of the five stages passes a DoD gate; a rejected stage triggers evidenced
rework with a bounded retry budget. \GoldEval scores the run against
the 120-case golden set with two grader types, and an Outcome-first
verdict accepts any valid trajectory that reaches the goal. Three-level
drill-down (task$\to$milestone$\to$step) of \eqref{eq:drilldown}
localizes failures and yields four verdict grades. Auditing is read-only and never modifies the original
trace, plan or artifacts.

The 30-day pre-asset observation of Section~\ref{sec:problem:objective}
supplies the baseline behaviour for this decomposition; its empty-rule
finding is the empirical origin of the deterministic judgement
requirement in Section~\ref{sec:evidence}.

\subsection{Cross-Team Transfer}
\label{sec:transfer}

In Domain B, a two-person non-original team with no prior harness code took over
in four days, raising solution-design stability from 50\% to 90\% with
post-release execution success above 90\% and a one-week capability
handover. In Domain C, one person completed integration and orchestration
debugging in seven days, reducing one day of manual stitching to five
hours under gate review and supporting a trial release. Accompanying
observations are 2--4 hours to 5--10 minutes for new-scenario setup and the
41st scenario costing approximately the same as the first. A large-scale
C-level deployment observation is reported separately in Appendix~F and is
not combined with the result-table measurements. These observations are
consistent with reuse of the same invariant set by teams that did not write
the assets.
Domain B accuracy $0.15\!\to\!0.6$ and Domain C
$\Fone$ thresholds $>0.6$/$>0.8$ are \emph{targets} specified by the
deployment.

\begin{table}[!htbp]
\centering
\caption{Cross-team transfer.}
\label{tab:transfer}
\small
\begin{tabularx}{\linewidth}{lYY}
\toprule
Domain & Effort & Outcome \\
\midrule
Domain A (first task force) & original team & 100\% completion; $-64\%$ latency; three of four recall targets reached \\
Domain B & 2 persons $\times$ 4 days, non-original & stability $50\%\!\to\!90\%$; execution success $>90\%$; one-week handover \\
Domain C & 1 person $\times$ 7 days & 1 day manual stitching $\to$ 5 h under gate review; V0.5 release \\
\bottomrule
\end{tabularx}
\par\vspace{2pt}
{\footnotesize\raggedright \textit{Note.} The Outcome column reports
measured quantities together with deployment milestones; C-level
observations are documented separately in Appendix~F.\par}
\end{table}

Table~\ref{tab:transfer} reports replication effort and outcome for the
two non-original teams; supplementary deployment-level observations are
documented in Appendix~F.

\FloatBarrier   % keep Section~\ref{sec:transfer}'s table before this subsection
\subsection{Evolution Dynamics}
\label{sec:evolution}

\begin{figure}[htbp]
\centering
\includegraphics[width=0.94\linewidth]{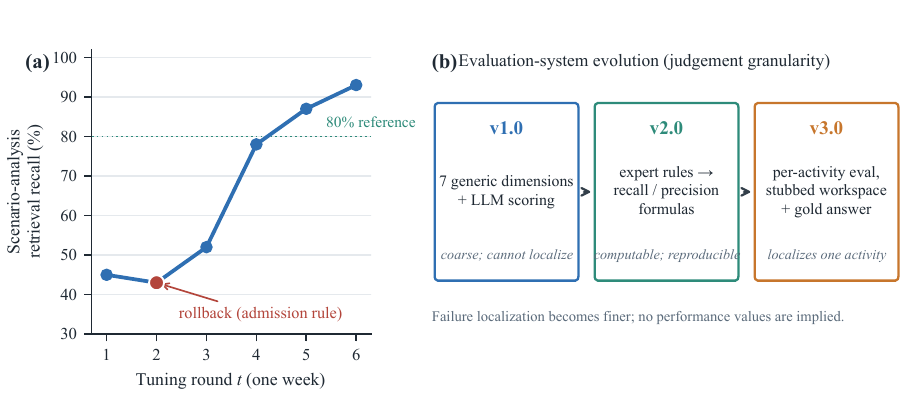}
\caption{Tuning dynamics and evaluation-system evolution. In (a) the six
rounds are measured on the scenario-analysis slice of the 120-case set; the
dotted 80\% line is a reference level of this retrieval-recall curve and is
not the four-activity target of Table~\ref{tab:main_results}(b). Panel (b)
shows the judgement granularity of the evaluation system; no performance
values are implied.}
\label{fig:tuning}
\end{figure}

Figure~\ref{fig:tuning} shows six rounds of tuning on the scenario-analysis
slice of the 120-case evaluation set: retrieval recall
$45\!\to\!43\!\to\!52\!\to\!78\!\to\!87\!\to\!93$. The curve is
non-monotonic; the round-2 drop is caught by the regression gate and rolled
back as prescribed by the regression-gated admission rule, after which recall rises
to 93\%. Panel (b) shows the evaluation system evolving from coarse LLM
scoring (v1.0), through computable expert-rule formulas (v2.0), to
per-activity evaluation with stubbed workspaces and the previous step's
reference answer (v3.0), progressively localizing failures to individual
activities. The measurements are organized along the
control-structure and knowledge-supply lines, using per-plane measurements,
the 30-day pre-asset observation of Section~\ref{sec:problem:objective}
and the rollback branch of the regression-gated admission rule. That observation
also supplies the empirical basis for deterministic machine judgement in
Section~\ref{sec:evidence}.

\subsection{Attribution Chain}
\label{sec:attribution}

Table~\ref{tab:attribution} links each invariant to the measured
end-to-end outcomes, together with the measurement slice of the
120-case evaluation set on which each value is reported.
The chain organizes measured associations: because the intervention changes
control structure, knowledge supply and evaluation jointly, it does not
identify the causal contribution of any single invariant. The slices are
views of one evaluation set rather than independent samples, so they are
reported separately without pooling and without a pooled significance
test; the chain therefore supports associations, not causal identification.

\begin{table}[!ht]
\centering
\caption{Attribution chain from invariants to measured end-to-end outcomes.
A = reproducible deterministic check; B = field measurement with a stated
sample size; C = descriptive deployment observation.}
\label{tab:attribution}
\small
\begin{tabularx}{\linewidth}{lYYl}
\toprule
Invariant / line & Intermediate observable & End-to-end outcome (measured) & Grade \\
\midrule
(2) Deterministic enforcement: DoD gates, externalized state, bounded rework & a latent failure becomes an observed, repairable event at $a_i$ & $\CR$ $50\%\!\to\!100\%$ and $L$ $440\!\to\!159$ min (120-case set, Table~\ref{tab:main_results}a) & B \\
(1) Externalized assets; generation--evaluation separation & control data inspectable before consumption; independent verdict & $\InvViol=0$ by construction (Appendix~A); about $+10\%$ accuracy (Table~\ref{tab:planes}, business) & A/B \\
Knowledge supply: $\Rrun$, $\varphi$, $\Rbuild$ & the evidence an activity needs is available & retrieval recall $56\%\!\to\!91.67\%$ (retrieval slice of the 120-case set); scenario recall $45\%\!\to\!93\%$ (120-case set, Table~\ref{tab:main_results}c) & B \\
(3) Read-only evidence: \GoldEval, ability/regression split, drill-down & failures localized to $M^{(2)}$ (step) & collaboration $\Fone$ $4\%\!\to\!61.4\%$, $L$ $49.6\!\to\!24.7$ min (Section~\ref{sec:main_results}); tuning $45\!\to\!43\!\to\!52\!\to\!78\!\to\!87\!\to\!93$ (Fig.~\ref{fig:tuning}) & B \\
Regression-gated versioning & assets reused without rewriting the harness & Domain B $2\times4$ days, stability $50\%\!\to\!90\%$; Domain C $1\times7$ days, $1$ day $\to$ $5$ h (Table~\ref{tab:transfer}) & B \\
\bottomrule
\end{tabularx}
\par\vspace{2pt}
{\footnotesize\raggedright \textit{Note.} All values are reported elsewhere
in this paper on the 120-case evaluation set; its activity, sub-task and
domain slices are reported separately. C-grade deployment observations are
documented in
Appendix~F.\par}
\end{table}

\paragraph{When the method does not apply.} The gates require activities
whose completion can be checked mechanically, a regression set that covers
the changed behaviour, and stable historical judgements for confidence
grading. In domains where acceptance criteria cannot be decided
mechanically, or where the regression set does not cover an activity, the
DoD gates and the admission rule cannot be applied as specified and the
human backstop of Section~\ref{sec:evidence} remains necessary. The
$\mathrm{pass@}3$ and $\mathrm{pass}^3$ figures are deployment targets,
not measured pass rates; the requirements-analysis recall of 78.08\% in
Table~\ref{tab:main_results}(b) remains below its 80\% target.

The three deployed domains share the invariant set of
Section~\ref{sec:method}; only the configuration constants are
re-instantiated. The chain above is the basis for the mechanism discussion
in this section. The controlled test of this decomposition is reported in
Section~\ref{sec:ablation}.

\subsection{Controlled Ablation}
\label{sec:ablation}

The attribution chain above organizes evaluation-set-level associations: the field
intervention changes control structure, knowledge supply and evaluation
together, so it does not identify the causal contribution of either failure
line on its own. We therefore ran two controlled ablation studies: an
\emph{additive} study that adds one line at a time to a fixed baseline
(B0--B4) and a \emph{subtractive} study that removes one line at a time
from the full system (A0--A6), followed by a threshold grid (A7) and a
cross-domain holdout (A8). Tables~\ref{tab:ablation_add}
and~\ref{tab:ablation_sub} report the arms, paired sample sizes and the
measured outcomes. The two studies are independent experiments with their own case
sets; they are compared with each other by sign and separation pattern only,
and are not pooled with the field evaluation set of Table~\ref{tab:main_results}.

\paragraph{Protocol.} Both studies held one backbone snapshot, one endpoint,
sampling parameters, component versions, golden sets and decision rules
fixed. Every arm used the same paired cases, randomized case and run order,
an independent work directory per run and $R=13$ independent repetitions,
with the run identifier, timestamp and component hash recorded. The
additive and subtractive studies each used $n=120$ paired cases;
the pairing unit is one case in one run. Content quality was produced by
\GoldEval separately from the execution report. Completion $\CR$ is
reported as a percentage with Wilson 95\% and Clopper--Pearson 95\%
intervals; the remaining quantities are means over the $R$ runs. Paired
comparisons use paired $t$ or Wilcoxon signed-rank with Holm correction
within each controlled study, reporting mean differences and effect sizes
rather than $p$-values alone. The $\Fone$ column is $2PR/(P+R)$ computed
from the reported means of $R$ and $P$; all three are case-level means over
the paired cases and the $R$ runs.

\begin{table}[!ht]
\centering
\caption{Additive controlled ablation ($n=120$, $R=13$). Completion is
reported as a percentage with its 95\% confidence interval computed by the
Wilson method and by the Clopper--Pearson exact method. $R$, $P$ and $\Fone$
are proportions in $[0,1]$. $L$ is the end-to-end latency in minutes. All
quantities are means over the $R=13$ runs. This study is not pooled with
Table~\ref{tab:main_results}.}
\label{tab:ablation_add}
\small
\begin{tabularx}{\linewidth}{lYrrrrrY}
\toprule
Arm & Component added & $\CR$ (\%) & \multicolumn{2}{c}{95\% CI} & $R$ / $P$ / $\Fone$ & $L$ (min) \\
\cmidrule(lr){4-5}
 & & & Wilson & CP & & \\
\midrule
B0 & none: monolithic agent, long prompt, no $K$ & 50.0 & [41.2, 58.8] & [40.7, 59.3] & 0.45 / 0.19 / 0.267 & 440 \\
B1 & DoD gates, externalized state, state machine & 75.0 & [66.6, 81.9] & [66.3, 82.5] & 0.55 / 0.24 / 0.334 & 250 \\
B2 & dual-engine retrieval and semantic alignment & 60.0 & [51.1, 68.3] & [50.7, 68.8] & 0.82 / 0.38 / 0.519 & 400 \\
B3 & both lines (Full) & 90.0 & [83.3, 94.2] & [83.2, 94.7] & 0.88 / 0.42 / 0.569 & 170 \\
B4 & longer prompt and more retries, compute-matched to B3 & 60.0 & [51.1, 68.3] & [50.7, 68.8] & 0.56 / 0.24 / 0.336 & 415 \\
\bottomrule
\end{tabularx}
\end{table}

\paragraph{Primary readouts and results.} One primary readout was fixed per
arm before execution. In the additive study, B1 moves completion and
latency ($+25$ pp on $\CR$ and $+0.067$ on $\Fone$), whereas B2 moves
content quality ($+10$ pp on $\CR$ and $+0.252$ on $\Fone$); B3 exceeds
the compute-matched B4 by $30$ pp. This is the \emph{double separation}
that the field attribution cannot isolate on its own: the control-structure
line governs completion and latency, and the knowledge-supply line governs
$R/\Fone$. The subtractive study shows the same signature with the opposite
sign: removing gates (A2) lowers $\CR$ by $22.5$ pp and raises $L$, while
removing retrieval (A5) lowers $\Fone$ by $0.198$ with only a $4.2$ pp
change in $\CR$. All three pre-specified sign conditions hold---the
control-structure condition ($\CR_{\mathrm{A2}}<\CR_{\mathrm{A1}}$ and
$L_{\mathrm{A2}}>L_{\mathrm{A1}}$), the knowledge-supply separation
($R/\Fone_{\mathrm{A5}}<R/\Fone_{\mathrm{A1}}$ with
$|\Delta\CR|\le5$ pp), and the exclusion condition
($\CR_{\mathrm{A1}}-\CR_{\mathrm{A6}}\ge15$ pp at matched token
budget)---and Holm correction was applied to the paired comparisons within
each controlled study. All arms are reported, including A4, whose small
$R/P$ gain reflects self-evaluation inflation rather than an improvement
and is not used to claim that generation--evaluation separation is
dispensable. The two studies are compared only by sign and separation
pattern, not in magnitude: they use different case sets and both contain
interaction terms ($+5$ pp and $11.7$ pp respectively), which are reported
but not summed.

\begin{table}[!ht]
\centering
\caption{Subtractive controlled ablation ($n=120$, $R=13$). Completion is
reported as a percentage with its 95\% confidence interval computed by the
Wilson method and by the Clopper--Pearson exact method. $R$, $P$ and $\Fone$
are proportions in $[0,1]$. $L$ is the end-to-end latency in minutes. All
quantities are means over the $R=13$ runs. A4 and A5 have the same
completion count but differ in content quality. This study is not pooled
with Table~\ref{tab:main_results}.}
\label{tab:ablation_sub}
\small
\begin{tabularx}{\linewidth}{lYrrrrrY}
\toprule
Arm & Component removed & $\CR$ (\%) & \multicolumn{2}{c}{95\% CI} & $R$ / $P$ / $\Fone$ & $L$ (min) \\
\cmidrule(lr){4-5}
 & & & Wilson & CP & & \\
\midrule
A0 & all: monolithic baseline & 50.0 & [41.2, 58.8] & [40.7, 59.3] & 0.45 / 0.19 / 0.267 & 440 \\
A1 & none (Full reference) & 88.3 & [81.4, 92.9] & [81.2, 93.5] & 0.88 / 0.42 / 0.569 & 159 \\
A2 & DoD gate and bounded rework & 65.8 & [57.0, 73.7] & [56.6, 74.2] & 0.83 / 0.39 / 0.531 & 388 \\
A3 & externalized state and checkpoint recovery & 70.8 & [62.2, 78.2] & [61.8, 78.8] & 0.82 / 0.38 / 0.519 & 340 \\
A4 & generation--evaluation separation & 84.2 & [76.6, 89.6] & [76.4, 90.2] & 0.90 / 0.43 / 0.582 & 174 \\
A5 & runtime retrieval and semantic alignment & 84.2 & [76.6, 89.6] & [76.4, 90.2] & 0.55 / 0.28 / 0.371 & 226 \\
A6-M & replaced by longer prompt and retries, token-matched & 61.7 & [52.7, 69.9] & [52.4, 70.4] & 0.56 / 0.24 / 0.336 & 415 \\
A6-L & as A6-M with $1.5\times$ token budget & 68.3 & [59.6, 76.0] & [59.2, 76.5] & 0.60 / 0.27 / 0.372 & 430 \\
\bottomrule
\end{tabularx}
\end{table}

\paragraph{Threshold and transfer arms.} A7 varies $Q$, $Q_{rt}$ and
$r_{\max}$ over $3\times3\times3=27$ cells; all three sign conditions hold
in every cell, so the conclusion is not sensitive to these thresholds. A8
re-uses the invariant set unchanged on a held-out domain and re-instantiates
only the six classes of per-domain configuration constants; no invariant
required rewriting, and completion stayed within $5$ pp of the in-domain
full system. These arms are implemented as independent configuration
branches of the reference runtime, with the thresholds injected explicitly.

\paragraph{Relation to the field measurements.} The $120/120$ completion
in Table~\ref{tab:main_results}(a) has a Wilson 95\% interval of
$[0.969,1.000]$ and a Clopper--Pearson 95\% interval of $[0.970,1.000]$;
both are derived from the reported count. No interval is assigned to the
remaining aggregate scores, which are averages over sentence-level
judgements rather than binomial counts. Figure~\ref{fig:error} is a purely
schematic product-form illustration and carries no measured data points.

%% file: conclusion.tex
\section{Conclusion}
\label{sec:conclusion}

We asked whether long-horizon business workflows can be made reliable
without changing the language model. Our answer is to keep the backbone
fixed and place the workflow definition, state and domain knowledge in
versioned control assets that a state machine, hooks and acceptance gates
enforce at run time. An independent auditor grades each trace, and asset
updates require a regression check and human approval. The released
implementation consists of an asset compiler, an execution runtime and a
read-only auditor; the four control planes, dual knowledge engines,
golden-set evaluation and confidence grading are configuration choices
of the deployment, not additional mechanisms.

The three invariants answer the challenges C1--C4 of
Section~\ref{sec:problem:objective}. Externalized assets make the inputs to
each activity inspectable before use; deterministic enforcement turns
failures into events that can be observed and repaired at the activity
where they occur; and a read-only audit limits the damage from a bad asset
version. On three deployed 200+ step domains, before/after measurements
show gains in completion, content quality and cross-team transfer under the
same fixed backbone, organized by the attribution chain of
Section~\ref{sec:attribution}. Each measurement slice is reported
separately, and the controlled ablation of Section~\ref{sec:ablation}
separates the two failure sources with a compute-matched exclusion control.

Two extensions remain: applying the same invariant set to further
long-horizon domains, and using the read-only audit to propose asset
revisions that still pass the regression check and human approval.

%% file: appendix.tex
\section{Implementation Details}

\paragraph{Component layout and versions.} Table~\ref{tab:components}
lists components and versions. The three released components are
implemented in \texttt{AgentGenesis-Code} (asset compilation),
\texttt{drive-2.3.4/drive} (declarative execution) and
\texttt{06-traceaudit-plugin} (read-only auditing). Task center, \GoldEval,
skill optimization and the dual knowledge engines are deployment-side
components described in the field material and component documents.

\begin{table}[h]
\centering
\caption{Component versions.}
\label{tab:components}
\small
\begin{tabularx}{\linewidth}{lYY}
\toprule
Component & Role & Version / maturity \\
\midrule
agent-genesis & asset production, six-phase pipeline, independent verifier & v0.17.10 \\
drive & declarative workflow, state machine, hooks & v2.3.4 (980 passed tests) \\
agent-evaluator (\GoldEval) & result/process evaluation, read-only audit & v1.0.0, no third-party deps \\
skill-optimizer & evaluation$\to$gap$\to$optimization loop, approval + rollback & 16 subagents \\
cosight-viewer (task center) & dual-channel capture, board, token cost, intervention & v1.2.7 \\
framework / smart-factory / cosight-bot & host / productization / preset experts & stable / skeleton / preset assets \\
\bottomrule
\end{tabularx}
\end{table}

\paragraph{Workflow schema and hooks.} \texttt{workflow.yaml} declares
phases, gates, skip conditions and evaluation levels; the artifact contract
(\texttt{path\_pattern}, \texttt{min\_count}, \texttt{min\_lines}) is
checked by the auditing plugin's structural checker. The state machine
infers legal transitions from the declared graph;
\texttt{hooks/state\_transition\_guard.py} blocks out-of-scope writes to
workflow control files; \texttt{Envelope.fail} is fail-closed; and
\texttt{verdict\_required} forces an explicit verdict at the terminal
state. High-risk actions pass a double HITL gate. The plugin manifest
states that no runtime engine is bundled. Table~\ref{tab:hooks} lists the
hooks and contracts.

\begin{table}[h]
\centering
\caption{Deterministic hooks and contracts referenced in
Section~\ref{sec:control}.}
\label{tab:hooks}
\small
\begin{tabularx}{\linewidth}{lY}
\toprule
Hook / contract & Enforced invariant \\
\midrule
\texttt{state\_transition\_guard.py} & only legal $\delta$ transitions; $\InvViol=0$ \\
artifact contract & \texttt{path\_pattern}, \texttt{min\_count}, \texttt{min\_lines} checked by the auditor \\
rework budget & $b_t\le3$ with attached evidence \\
regression gate & admission only after regression check and human approval \\
read-only audit & never modifies $\tau$, plan or artifacts \\
\bottomrule
\end{tabularx}
\end{table}

\noindent Reproduction entry points are \texttt{bash run-e2e.sh},
\texttt{python3 scripts/run\_e2e.py --task-file ...},
\texttt{./bin/traceaudit run|post-execute|validate} and
\texttt{npm run smoke}. The auditing plugin does not auto-subscribe to
completion events; an adapter calls \texttt{post-execute}. The reference
implementation is released at
\url{https://github.com/ZTE-AICloud/Co-Sight/tree/master/Co-Sight%20Harness}.

\subsection{Invariant Conformance Check}
The released reference state machine declares four states
(\textsc{Idle}/\textsc{Processing}/\textsc{Review}/\textsc{Done}) and
five legal transitions. We injected all 16 ordered state pairs into
\texttt{scripts/state\_machine.py}, with a fresh audit log per case:
the 11 pairs outside the declared transition set were each rejected with
\texttt{TransitionError}, the 5 declared pairs were each admitted, and
re-reading the append-only audit log returned identical entries. This
conformance check establishes that the fail-closed transition guard and
the append-only log behave as specified in the state-machine invariant.
The check is reproducible from \texttt{scripts/state\_machine.py} in the
released asset compiler (Appendix~A).

\section{Algorithm Pseudocode}

Algorithm~\ref{alg:loop} in Section~\ref{sec:formalization} expands into
the five phases listed there. Phase~I (\texttt{Diagnose} and
\texttt{Compile}) grades each activity on five dimensions, emits a
priority order and subagent split, then runs intent routing,
clarification, analysis, plan-mode approval, assembly, static
verification and independent runtime evaluation; the compilation gate
then admits $K$. Phase~II (\texttt{Execute}) loads $K$,
advances the state machine, fires pre- and post-execution hooks and DoD
gates, enforces the artifact contract and default-deny write allowlist,
and returns $\tau$ with an append-only event log; dependency-ordered
waves are reconciled at a barrier and high-risk actions are held at a
double HITL gate. Phase~III (\texttt{Audit}) maps $\tau$ to $z$ through
four audit layers and computes
\eqref{eq:evidence-cap}--\eqref{eq:drilldown}, returning drill-down
metrics and an evidence record without modifying the trace. Phase~IV
(\texttt{Evaluate}) scores LLM-judged and deterministic items separately
in \GoldEval, reports ability and regression outcomes over disjoint
sets, and formulates a repair proposal from localized gaps. Phase~V
(\texttt{Merge}) admits $K'$ only when the regression-gated admission rule holds
and human approval is recorded; otherwise it rolls back to $K_t$. The
merge is read-only with respect to $\tau$ throughout.

\section{Instructions and Rubrics}

The compilation prompt family includes intent clarification, phase
analysis, plan-mode approval, assembly, static verification and runtime
evaluation templates; the independent verifier receives the artifact but
not the generator's rationale. \GoldEval decision rules for LLM-judged
items split reference and candidate answers into sentences, classify each
as hit / extra / missing and compute $R$, $P$, $\Fone$; deterministic items
use exact or containment matching. The six audit dimensions are weighted
by $w$, with evidence caps \eqref{eq:evidence-cap} and red-line minima
\eqref{eq:audit-total}. Golden-set examples are drawn from the 120-case
evaluation set; each carries a reference answer and a decision rule. Ability and
regression thresholds are $\mathrm{pass@}3\ge90\%$ and
$\mathrm{pass}^3=100\%$; these are deployment targets, not measured pass
rates.

\section{Trajectory Examples}

A representative fragment illustrates a DoD rejection and evidenced
rework. An activity produces an artifact whose contract requires a minimum
line count; the hook rejects it, attaches the failing contract field to the
event log, and returns the activity to analysis with $b_t=1$. The second
attempt satisfies the contract and passes the deterministic grader, while
the \GoldEval LLM grader records hit / extra / missing sentence classes.
Three-level drill-down attributes the failure to $M^{(2)}$ (step) rather
than $M^{(0)}$ (task). The auditing layer reads $\tau$ only; the original
plan and artifacts are unchanged; the case is documented at the level of
verdict grades and drill-down attribution.

\section{Control Asset Artifacts}

Control assets fall into four classes: goals and plans, state and
checkpoints, intermediate artifacts, and distilled experience. Examples
include a project brief, a roadmap, an execution plan, a locked state file
and an acceptance specification. Each asset is versioned ($K_t\to
K_{t+1}$) and every merge requires a regression check and human
approval. Configuration changes do not require code changes, which is the
mechanism behind cross-team transfer in Section~\ref{sec:transfer}.

\section{Supplementary Results (Field Observations)}

Table~\ref{tab:observations} collects C-level deployment observations:
non-controlled, desensitized records documented separately from the
result-table measurements of Table~\ref{tab:main_results}. We observe these
values at the deployment level; they are descriptive and are kept separate
from the controlled evaluation-set results.

\begin{table}[h]
\centering
\caption{C-level field observations (non-controlled, descriptive).}
\label{tab:observations}
\small
\begin{tabularx}{\linewidth}{lYYlY}
\toprule
Indicator & Before & After & Sample / window & Source \\
\midrule
Aggregate task success & 60\% & 90\% & 2000 persons & product case (desensitized) \\
New-scenario setup & 2--4 h & 5--10 min & not specified & product case \\
Evaluation system & \multicolumn{2}{l}{v1.0 $\to$ v2.0 $\to$ v3.0} & coarse $\to$ computable $\to$ per-activity & product case \\
\bottomrule
\end{tabularx}
\par\vspace{2pt}
{\footnotesize\raggedright \textit{Note.} Non-controlled, desensitized
field observations; not comparable to Table~\ref{tab:main_results}.
Percentages are aggregate values as reported in the field material.
Domain-B stability and the tuning curve are evaluation-set measurements,
reported in Tables~\ref{tab:transfer} and~\ref{tab:attribution}.\par}
\end{table}

\section{Data Card}

Table~\ref{tab:datacard} records the composition of the 120-case
evaluation set and its measurement boundaries. The four-stage $\Fone$
values (62.66/60.50/62.27/56.4) and
Table~\ref{tab:main_results}(b) recall values are reported separately.
Percentages in the result tables are measured on this evaluation set;
the descriptive C-grade observations in Appendix~F are excluded.

The measurement conditions are fixed by the deployment. The backbone is
reached through an external API endpoint; the released components contain
no model weights, and one backbone serves both the before and after
conditions. The runtime is a standard agent host with the three control
plugins installed; component versions and reproduction entry points are
listed in Appendix~A. \GoldEval decision rules split LLM-judged items
into sentences, classify them as hit / extra / missing and compute $R$,
$P$ and $\Fone$; deterministic items use exact or containment matching,
and a case passes when its score clears the rule threshold. Verdict
grades $G_0$--$G_3$ denote crash, generation failure, quality failure
and acceptance. Ability and regression sets are disjoint
($\mathrm{pass@}3\ge90\%$ and $\mathrm{pass}^3=100\%$ are deployment
thresholds). Confidence is graded in three levels---human confirmation,
LLM verification with sampling, and deterministic checks without an LLM
call---and a human may exit the loop when the DoD is mechanically
decidable, the regression set has sufficient coverage and historical
judgements are stable. The 120-case evaluation set is partitioned into
activity, sub-task and domain slices that are reported separately; the six
tuning rounds span one week and the pre-asset behaviour observation spans
30 days.

\begin{table}[h]
\centering
\caption{Data card: composition of the 120-case evaluation set.}
\label{tab:datacard}
\small
\begin{tabularx}{\linewidth}{lYY}
\toprule
Item & Value & Note \\
\midrule
Evaluation set & 120 end-to-end cases & reference answers and decision rules, disjoint ability / regression subsets \\
Activity slices & four full-flow activities & requirements / system-solution / subsystem / component design \\
Retrieval / scenario slices & three retrieval sub-tasks & reported separately, not pooled \\
Four-stage $\Fone$ & 62.66 / 60.50 / 62.27 / 56.4 & aggregate over the same evaluation set \\
drive tests & 980 passed & CHANGELOG \\
Counts & 120 golden cases; 34/34 templates; 90/90 unit tests; 102 YAML cases & measured in code \\
\midrule
\multicolumn{3}{l}{Thresholds (not results): $\mathrm{pass@}3\ge90\%$, $\mathrm{pass}^3=100\%$, Domain B accuracy target 0.6, Domain C $\Fone$ targets $>0.6$/$>0.8$.} \\
\multicolumn{3}{l}{Percentages in the result tables are measured on the 120-case evaluation set; the C-grade observations of Appendix~F are excluded.} \\
\multicolumn{3}{l}{Illustrative only: the $0.95^{10}\approx60\%$ curve depicts the product-form mechanism.} \\
\bottomrule
\end{tabularx}
\end{table}

%% file: statements.tex
\section*{Statements}

\paragraph{AI Use Statement.} Generative AI tools were used to assist with
the drafting and language editing of this manuscript and with the
preparation of figure prompts and plotting scripts. All technical claims,
measurements and code descriptions were checked against the released
implementation and the accompanying field material by the authors.

\paragraph{Reproducibility Statement.} The three released components---the
asset compiler, the declarative execution runtime and the read-only
trajectory auditor---implement the compilation, deterministic enforcement
and read-only auditing parts of the method: the frozen symbols and
invariants of Section~\ref{sec:problem}, Phases~I--III of
Algorithm~\ref{alg:loop}, and the workflow schema, state machine, hooks,
DoD gates and read-only auditing of Section~\ref{sec:method}. Asset
admission follows the regression-gated admission rule with human approval.
Repository names, versions
and reproduction
commands are listed in Appendix~A; the components can be cloned and
exercised with those commands. Evaluation-set composition, slice
distinctions and measurement conditions for the result tables are recorded
in the data card
(Appendix~G). A single backbone snapshot, reached through the same
OpenAI-compatible endpoint, served both the before and after conditions;
the model identifier, snapshot date, endpoint type and sampling parameters
(temperature, top-p, maximum tokens, seed) were held fixed and are recorded
with the deployment configuration. Deployment-side components (task
center, \GoldEval, skill optimization, knowledge engines) are configuration
and engineering practice
described in Section~\ref{sec:method}. No result in this paper depends on
proprietary model weights.

\paragraph{Replication protocol.} Replications hold the endpoint, model
version, sampling parameters, component versions, golden sets and decision
rules fixed, repeat each measurement slice independently, and record, for
every measured entry of Table~\ref{tab:main_results}, the mean over the
repetitions together with the run-level records from which dispersion can
be recomputed; the result tables report these means, and completion is
additionally reported with Wilson and Clopper--Pearson intervals.
Process-level evidence follows the same
protocol with per-stage
instrumentation of the process indicators defined in
Section~\ref{sec:problem:metrics}---$\InvViol$, $\HI$, $\LAop@\ell$,
$\Reg$ and $S$---recorded on the disjoint ability and regression sets.
Cross-model replication holds the control
assets fixed and varies only the endpoint. External baselines require a
shared task set and grading settings; the deployed pre-intervention
system is the reference for the before column. Appendix~G records the
measurement boundaries these replications must preserve. The controlled
ablation of Section~\ref{sec:ablation} follows the same protocol; its two
studies use independent configuration branches of the released runtime
and are reported separately from the field evaluation set.

\paragraph{Licence note.} \texttt{PRIMEarxiv.sty} is derived from the
\texttt{arxiv-style} project by George Kour and adapted by
Moulay A. Akhloufi for the PRIME AI group. The original licence notice is
retained in the source header of the file. Local modifications are limited
to removing the third-party lab header and disabling verbose geometry
logging. Licence statements for the submission and for the released
components are recorded separately.

%% file: references.bib
@article{madatha2026controlplane,
  title   = {A Deterministic Control Plane for {LLM} Coding Agents},
  author  = {Madatha, Padmaraj},
  journal = {arXiv preprint arXiv:2606.26924},
  year    = {2026},
  url     = {https://arxiv.org/abs/2606.26924},
  note    = {45 pages, 9 figures, 13 tables; dataset and reproduction scripts at Zenodo}
}

@article{fournier2026processharness,
  title   = {A Process Harness for Uplifting Legacy Workflows to Agentic {BPM}: Design and Realization in {CUGA} {FLO}},
  author  = {Fournier, Fabiana and Limonad, Lior},
  journal = {arXiv preprint arXiv:2606.27188},
  year    = {2026},
  url     = {https://arxiv.org/abs/2606.27188}
}

@inproceedings{zhao2026evoml,
  title     = {Validation-Gated {ML} Pipeline Synthesis with Expert-Grounded Self-Evolving Skills},
  author    = {Zhao, Zixiao and Sun, Jing and Hou, Zhe and Li, Mengze and Liu, Qian and Cai, Cheng-Hao and Zhang, Zijian and Wang, Xin and Dobbie, Gillian and Dong, Jin Song},
  booktitle = {Proceedings of the 49th International Conference on Software Engineering ({ICSE})},
  year      = {2026},
  url       = {https://arxiv.org/abs/2611.04180}
}

@inproceedings{orzechowski2026sla,
  title     = {When the Consumer Is an Agent: Execution-Grounded Service-Level Agreements},
  author    = {Orzechowski, Michal and Slota, Renata G. and Kitowski, Jacek},
  booktitle = {Proceedings of the 24th International Conference on Service-Oriented Computing ({ICSOC} 2026)},
  year      = {2026},
  publisher = {Springer},
  url       = {https://arxiv.org/abs/2611.04182}
}

@inproceedings{gupta2026agenttrustci,
  title     = {{AgentTrustCI}: Build-Time Policy, Grounding, and Trajectory Tests for {Java} {LLM} Agents},
  author    = {Gupta, Sachin},
  booktitle = {Proceedings of the 2026 International Conference on Computer and Applications ({ICCA})},
  year      = {2026},
  url       = {https://arxiv.org/abs/2611.06674}
}

@article{shan2026statemirror,
  title   = {{StateMirror}: Pre-Execution Detection of Hallucinated Tool Calls in Coding Agents},
  author  = {Shan, Shiwen and Huo, Yintong and Wu, Zhiying and Su, Yuxin},
  journal = {arXiv preprint arXiv:2610.32648},
  year    = {2026},
  url     = {https://arxiv.org/abs/2610.32648}
}

@article{pan2026harnesspro,
  title   = {{HarnessPro}: Recursive Harness Self-Improvement with Pareto Optimization for Performance-Cost Trade-offs},
  author  = {Pan, Zhuoshi and Feng, ZhiYuan and Cen, Yukuo and Lin, Honglin and Lu, Junru and Chen, Zhongxia and Yang, Shuo and Zhao, H. Vicky and Tang, Jie},
  journal = {arXiv preprint arXiv:2610.26770},
  year    = {2026},
  url     = {https://arxiv.org/abs/2610.26770},
  note    = {Under review; 41 pages, 4 figures, 4 tables}
}

@article{huyut2026harnessmodel,
  title   = {The Harness Is the Model: Agent Scaffolding Changes a Fixed {LLM}'s Compliance with Banking Conduct Rules, Unevenly Across Models},
  author  = {Huyut, Mehmet Murat},
  journal = {arXiv preprint arXiv:2611.06116},
  year    = {2026},
  url     = {https://arxiv.org/abs/2611.06116}
}

@inproceedings{gefen2026cleo,
  title     = {Optimize the Harness, Not the Model: Fitting a Minimal Agent to a Self-Hostable {LLM} for Investigative Analysis},
  author    = {Nahum Gefen, Shai and Chassida, Vered and Basin, Dmitry and Feigenblat, Guy},
  booktitle = {Proceedings of the 31st {ACM} International Conference on Information and Knowledge Management ({CIKM} 2026), Industry Day},
  year      = {2026},
  address   = {Rome, Italy},
  doi       = {10.1145/3799682.3839256},
  url       = {https://arxiv.org/abs/2611.03434}
}

@inproceedings{zhang2025aflow,
  title     = {{AFlow}: Automating Agentic Workflow Generation},
  author    = {Zhang, Jiayi and Xiang, Jinyu and Yu, Zhaoyang and Teng, Fengwei and Chen, Xionghui and Chen, Jiaqi and Zhuge, Mingchen and Cheng, Xin and Hong, Sirui and Wang, Jinlin and others},
  booktitle = {The Thirteenth International Conference on Learning Representations ({ICLR})},
  year      = {2025},
  url       = {https://arxiv.org/abs/2410.10762}
}

@inproceedings{zhang2026dgm,
  title     = {Darwin Godel Machine: Open-Ended Evolution of Self-Improving Agents},
  author    = {Zhang, Jenny and Hu, Shengran and Lu, Cong and Lange, Robert and Clune, Jeff},
  booktitle = {The Fourteenth International Conference on Learning Representations ({ICLR})},
  year      = {2026},
  url       = {https://arxiv.org/abs/2505.22954}
}

@inproceedings{pan2026rho,
  title     = {Evolving Agents in the Dark: Retrospective Harness Optimization via Self-Preference},
  author    = {Pan, Wenbo and Liu, Shujie and Lin, Chin-Yew and Zeng, Jingying and Tang, Xianfeng and Zhou, Xiangyang and Lu, Yan and Jia, Xiaohua},
  booktitle = {Findings of the Association for Computational Linguistics: EMNLP 2026},
  year      = {2026},
  url       = {https://arxiv.org/abs/2606.05922}
}

@article{chen2026purposeir,
  title   = {Purpose {IR}: Compiling User Purpose for Privacy-Preserving Agent Workflow State},
  author  = {Chen, Dongsheng and Zhao, Xiangyu and Ma, Lei and Yao, Xin and Wei, Xuetao},
  journal = {arXiv preprint arXiv:2610.16414},
  year    = {2026},
  url     = {https://arxiv.org/abs/2610.16414}
}

@article{fang2026toolqp,
  title   = {Beyond Single-Shot: Multi-step Tool Retrieval via Query Planning},
  author  = {Fang, Wei and Glass, James},
  journal = {arXiv preprint arXiv:2601.07782},
  year    = {2026},
  url     = {https://arxiv.org/abs/2601.07782}
}

@inproceedings{ouyang2026reasoningbank,
  title     = {{ReasoningBank}: Scaling Agent Self-Evolving with Reasoning Memory},
  author    = {Ouyang, Siru and Yan, Jun and Hsu, I-Hung and Chen, Yanfei and Jiang, Ke and Wang, Zifeng and Han, Rujun and Le, Long and Daruki, Samira and Tang, Xiangru and others},
  booktitle = {The Fourteenth International Conference on Learning Representations ({ICLR})},
  year      = {2026},
  url       = {https://arxiv.org/abs/2509.25140}
}

@inproceedings{suzgun2026dynamiccheatsheet,
  title     = {Dynamic Cheatsheet: Test-Time Learning with Adaptive Memory},
  author    = {Suzgun, Mirac and Yuksekgonul, Mert and Bianchi, Federico and Jurafsky, Dan and Zou, James},
  booktitle = {Proceedings of the 19th Conference of the European Chapter of the Association for Computational Linguistics (Volume 1: Long Papers)},
  month     = mar,
  year      = {2026},
  address   = {Rabat, Morocco},
  publisher = {Association for Computational Linguistics},
  pages     = {7080--7106},
  doi       = {10.18653/v1/2026.eacl-long.333},
  url       = {https://aclanthology.org/2026.eacl-long.333/}
}

@inproceedings{pan2026contextpilot,
  title     = {{ContextPilot}: Teaching Agents for Proactive Context Management via Fine-grained {RL}},
  author    = {Pan, Zhuoshi and Pei, Qizhi and Lu, Junru and Lin, Honglin and Zhao, H. Vicky and Yin, Di and Sun, Xing},
  booktitle = {Proceedings of the 2026 Conference on Empirical Methods in Natural Language Processing},
  month     = nov,
  year      = {2026},
  address   = {Budapest, Hungary},
  publisher = {Association for Computational Linguistics},
  url       = {https://arxiv.org/abs/2608.28476}
}

@inproceedings{yehudai2026survey,
  title     = {A Survey on Evaluation of {LLM}-based Agents},
  author    = {Yehudai, Asaf and Eden, Lilach and Li, Alan and Uziel, Guy and Zhao, Yilun and Bar-Haim, Roy and Cohan, Arman and Shmueli-Scheuer, Michal},
  booktitle = {Findings of the Association for Computational Linguistics: {ACL} 2026},
  month     = jul,
  year      = {2026},
  address   = {San Diego, California, United States},
  publisher = {Association for Computational Linguistics},
  pages     = {26690--26714},
  doi       = {10.18653/v1/2026.findings-acl.1330},
  url       = {https://aclanthology.org/2026.findings-acl.1330/}
}

@inproceedings{guo2026agenteval,
  title     = {{AgentEval}: {DAG}-Structured Step-Level Evaluation for Agentic Workflows with Error Propagation Tracking},
  author    = {Guo, Dongxin and Wu, Jikun and Yiu, Siu Ming},
  booktitle = {Proceedings of the 64th Annual Meeting of the Association for Computational Linguistics ({ACL} 2026), Industry Track},
  year      = {2026},
  url       = {https://arxiv.org/abs/2604.23581}
}

@article{charoosaei2026llmregtests,
  title   = {On the Risks of using {LLM}-Generated Tests for Regression Testing},
  author  = {Charoosaei, Mohammadali and Richter, Cedric and Papadakis, Mike},
  journal = {arXiv preprint arXiv:2610.11835},
  year    = {2026},
  url     = {https://arxiv.org/abs/2610.11835}
}

@misc{grace2026demystifying,
  title        = {Demystifying Evals for {AI} Agents},
  author       = {Grace, Mikaela and Hadfield, Jeremy and Olivares, Rodrigo and De Jonghe, Jiri},
  howpublished = {Anthropic Engineering},
  year         = {2026},
  month        = jan,
  note         = {Published January 9, 2026},
  url          = {https://www.anthropic.com/engineering/demystifying-evals-for-ai-agents}
}

@inproceedings{abichandani2026rsd,
  title     = {Recursive Semantic Divergence for {LLM} Agent Consistency},
  author    = {Abichandani, Harshavardhan and Chong, Penny and Ghosh, Atin and Dahlmeier, Daniel},
  booktitle = {Advances in Neural Information Processing Systems ({NeurIPS})},
  year      = {2026},
  url       = {https://arxiv.org/abs/2610.33012}
}

@inproceedings{shu2026escrow,
  title     = {{ESCROW}: Guarded and Dual-Objective Continual Maintenance for Agents in Policy-Governed Enterprise Workflows},
  author    = {Shu, Ruoqi and Dan, Chen and Wang, Xuhui and Xu, Tianhua and Luo, Mengxi and Mai, Yanming and Wan, Bo},
  booktitle = {CLEA Workshop at Advances in Neural Information Processing Systems ({NeurIPS}) 2026},
  year      = {2026},
  url       = {https://arxiv.org/abs/2608.01772}
}

@inproceedings{cao2026longbanana,
  title     = {{LongBanana}: An Expert-Verified Benchmark for Long-Context Multi-Reference Image Synthesis},
  author    = {Cao, Haoxiang and Zhang, Yuxuan and Du, Penghui and Li, Bo and Cao, Jiajiong and Fan, Yingying and Wu, Zimeng and Luo, Yifu and Zheng, Mingzhe and Miao, Junwen and others},
  booktitle = {Advances in Neural Information Processing Systems ({NeurIPS}) 2026, Datasets and Benchmarks Track},
  year      = {2026},
  url       = {https://arxiv.org/abs/2611.06856}
}

@inproceedings{liu2026bats,
  title     = {Budget-Aware Tool Use Enables Effective Agent Scaling},
  author    = {Liu, Tengxiao and Wang, Zifeng and Miao, Jin and Hsu, I-Hung and Yan, Jun and Chen, Jiefeng and Han, Rujun and Xu, Fangyuan and Chen, Yanfei and Jiang, Ke and others},
  booktitle = {Conference on Language Modeling ({COLM})},
  year      = {2026},
  url       = {https://arxiv.org/abs/2511.17006}
}

@article{breckenridge2026ihitl,
  title   = {{I-HITL}: Inalienable Human-in-the-Loop},
  author  = {Breckenridge, Samuel and Vilardell, Dani and Wang, Wenhao and Zhang, Fan and Juels, Ari},
  journal = {arXiv preprint arXiv:2610.35835},
  year    = {2026},
  url     = {https://arxiv.org/abs/2610.35835}
}

@inproceedings{cheng2026threepillar,
  title     = {A Three-Pillar Governance Framework for Safe and Trustworthy {AI} Agents},
  author    = {Cheng, Edward C. and Cheng, Jeshua and Siu, Alice},
  booktitle = {IEEE International Conference on Agentic, Generative, and Cognitive AI Systems ({AGCS} 2026)},
  year      = {2026},
  url       = {https://arxiv.org/abs/2610.30445}
}

@article{liu2026deliverbench,
  title   = {{DeliverBench}: Benchmarking {AI} Agents on Delivering Realistic Multi-Stage Professional Projects},
  author  = {Liu, Yumou and Chen, Mingju and Wen, Zhoufutu and Yang, Qianyu and Wang, Zijie and Chen, Sihan and Liu, Yeqian and Yang, Chenghao and Duan, Tiliang and Cai, Qiyu and others},
  journal = {arXiv preprint arXiv:2611.03412},
  year    = {2026},
  url     = {https://arxiv.org/abs/2611.03412}
}

@article{wu2026mattoolbench,
  title   = {{MatToolBench}: Benchmarking Multimodal Agents in Real-World Materials Science Workflows},
  author  = {Wu, Mei and Xie, Rui and Zhang, Runyu and Li, Yuqiang and Fu, Tianfan and Chen, Bo and Yu, Kai and Chen, Xin and Chen, Lu},
  journal = {arXiv preprint arXiv:2609.37053},
  year    = {2026},
  url     = {https://arxiv.org/abs/2609.37053}
}

@inproceedings{wei2026slate,
  title     = {{SLATE}: A Large-Scale Benchmark for Long-Horizon {API} Selection in E-commerce Tool Use},
  author    = {Wei, Rongzhe and Shi, Ge and Cheng, Min and Ghosh, Sarthak and Gorde, Vaibhav and Akoglu, Leman and Li, Pan and Zhang, Na},
  booktitle = {NeurIPS 2026 Workshop on Agentic AI Benchmarks and Applications for Enterprise Tasks ({AABA4ET})},
  year      = {2026},
  url       = {https://arxiv.org/abs/2611.03982}
}

@inproceedings{deng2026swebenchpro,
  title     = {{SWE-Bench Pro}: Can {AI} Agents Solve Long-Horizon Software Engineering Tasks?},
  author    = {Deng, Xiang and Da, Jeff and Pan, Edwin and He, Yannis Yiming and Ide, Charles and Garg, Kanak and Lauffer, Niklas and Park, Andrew and Pasari, Nitin and Rane, Chetan and others},
  booktitle = {Proceedings of the 43rd International Conference on Machine Learning ({ICML})},
  year      = {2026},
  url       = {https://arxiv.org/abs/2509.16941}
}

@inproceedings{erol2026costofpass,
  title     = {Cost-of-Pass: An Economic Framework for Evaluating Language Models},
  author    = {Erol, Mehmet Hamza and El, Batu and Suzgun, Mirac and Yuksekgonul, Mert and Zou, James},
  booktitle = {The Fourteenth International Conference on Learning Representations ({ICLR})},
  year      = {2026},
  url       = {https://arxiv.org/abs/2504.13359}
}
